\documentclass[10pt,twocolumn,letterpaper]{article}

\usepackage{cvpr}
\usepackage{times}
\usepackage{epsfig}
\usepackage{graphicx}
\usepackage{amsmath}
\usepackage{amssymb}
\usepackage{array}
\usepackage{multirow}
\usepackage{cite}
\usepackage[keeplastbox]{flushend}
\usepackage[normalem]{ulem}
\usepackage{fancyhdr}

\newcolumntype{?}[1]{!{\vrule width #1}}
\newcommand{\0}{\phantom{0}}

\newcommand{\hide}[1]{}

\newcommand{\Figs}{figs_eps}
\newcommand{\Figssupp}{figs_supp_eps}

\def\TPdel#1{}

\newcommand{\PAR}[1]{\vskip4pt \noindent{\bf #1~}}
\let\oldenumerate\enumerate
\renewcommand{\enumerate}{
\oldenumerate
\setlength{\itemsep}{1.2pt}
\setlength{\parskip}{0pt}
\setlength{\parsep}{0pt}
}

\let\olditemize\itemize
\renewcommand{\itemize}{
\olditemize
\setlength{\itemsep}{1.2pt}
\setlength{\parskip}{0pt}
\setlength{\parsep}{0pt}
}

\usepackage[pagebackref=true,breaklinks=true,letterpaper=true,colorlinks,bookmarks=false]{hyperref}

\cvprfinalcopy 

\def\cvprPaperID{853} 

\begin{document}

\title{Benchmarking 6DOF Outdoor Visual Localization in Changing Conditions}

\author{Torsten Sattler$^1$ \quad Will Maddern$^2$ \quad Carl Toft$^3$ \quad Akihiko Torii$^4$ \quad Lars Hammarstrand$^3$\\
Erik Stenborg$^3$ \quad Daniel Safari$^{4,5}$ \quad Masatoshi Okutomi$^4$ \quad Marc Pollefeys$^{1,6}$\\ %
\quad Josef Sivic$^{7,8}$ \quad 
Fredrik  Kahl$^{3,9}$ \quad Tomas Pajdla$^8$\\ 
$^1$Department of Computer Science, ETH Z\"{u}rich
\quad %
$^2$Oxford Robotics Institute, University of Oxford\\
$^3$Department of Electrical Engineering, Chalmers University of Technology \quad $^6$Microsoft\\ 
$^4$Tokyo Institute of Technology \quad 
$^5$Technical University of Denmark \quad  
$^7$Inria\thanks{WILLOW project, Departement d'Informatique de l'\'Ecole Normale Sup\'erieure, ENS/INRIA/CNRS UMR 8548, PSL Research University.}\\
$^8$CIIRC, CTU in Prague\thanks{CIIRC - Czech Institute of Informatics, Robotics, and Cybernetics, Czech Technical University in Prague} 
 \quad $^9$Centre for Mathematical Sciences, Lund University
}

\maketitle
\rhead{\small \textit{This is an extended version of a paper accepted for publication at CVPR 2018.  Main part of the paper: \textcopyright 2018 IEEE}}
\thispagestyle{fancy} 
\begin{abstract}
\vspace{-6pt}
\noindent Visual localization enables autonomous vehicles to navigate in their surroundings and augmented reality applications to link virtual to real worlds. Practical visual localization approaches need to be robust to a wide variety of viewing condition, including day-night changes, as well as weather and seasonal variations, while providing highly accurate 6 degree-of-freedom (6DOF) camera pose estimates. In this paper, we introduce the first benchmark datasets specifically designed for analyzing the impact of such factors on visual localization. Using carefully created ground truth poses for query images taken under a wide variety of conditions, we evaluate the impact of various factors on 6DOF camera pose estimation accuracy through extensive experiments with state-of-the-art localization approaches. Based on our results, we draw conclusions about the difficulty of different conditions, showing that long-term localization is far from solved, and  propose promising avenues for future work, including sequence-based localization approaches and the need for better local features. Our benchmark is available at \url{visuallocalization.net}.
\end{abstract}

\vspace{-12pt}
\section{Introduction}
\noindent Estimating the 6DOF camera pose of an image with respect to a 3D scene model is key for visual navigation of autonomous vehicles and augmented/mixed reality devices. 
Solutions to this \emph{visual localization} problem can also be used to ``close loops'' in the context of SLAM or to register images to Structure-from-Motion (SfM) reconstructions.

\begin{figure}[t]
  \centering
  \includegraphics[width=1\linewidth]{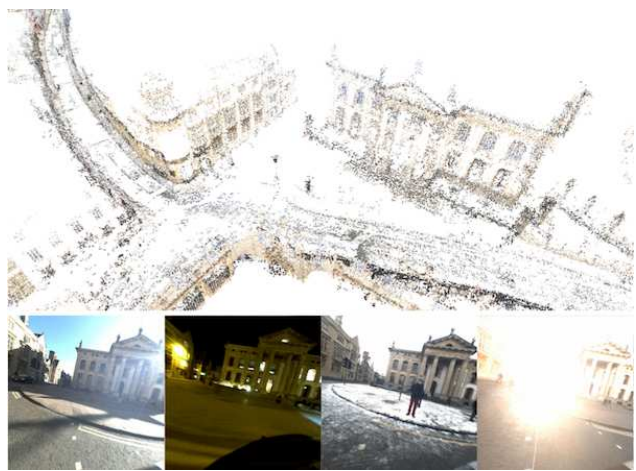}%
  \caption{Visual localization in changing urban conditions.
  We present three new datasets, \emph{Aachen Day-Night}, \emph{RobotCar Seasons} (shown) and \emph{CMU Seasons} for evaluating 6DOF localization against a prior 3D map (top) using registered query images taken from a wide variety of conditions (bottom), including day-night variation, weather, and seasonal changes over long periods of time.
  }
  \label{fig:robotcar-projection-1}
\end{figure}

Work on 3D structure-based visual localization has focused on increasing efficiency~\cite{Li10ECCV,Sattler17PAMI,Larsson16,Lynen15RSS,Taira2018CVPR}, improving scalability 
and robustness to ambiguous structures~\cite{Zeisl15ICCV,Svarm17PAMI,Sattler-ICCV-2015,Li12ECCV}, reducing memory requirements~\cite{Sattler-ICCV-2015,Li10ECCV,Cao14CVPR}, and more flexible scene representations~\cite{Sattler2017CVPR}. 
All these methods utilize local features to establish 2D-3D matches. 
These correspondences are in turn used to estimate the camera pose. 
This data association stage is critical as pose estimation fails without sufficiently many correct matches. 
There is a well-known trade-off between discriminative power and invariance for local descriptors. 
Thus, existing localization approaches will only find enough matches if both the query images and the images used to construct the 3D scene model are taken under similar viewing conditions. 

Capturing a scene under all viewing conditions is prohibitive. Thus, the assumption that all relevant conditions are covered is too restrictive in practice. 
It is more realistic to expect that images of a scene are taken under a single or a few conditions. 
To be practically relevant, \eg, for life-long localization for self-driving cars, visual localization algorithms need to be robust under varying conditions (\cf Fig.~\ref{fig:robotcar-projection-1}). 
Yet, no work in the literature actually measures the impact of varying conditions on 6DOF pose accuracy.

One reason for the lack of work on visual localization under varying conditions is a lack of suitable benchmark datasets. 
The standard approach for obtaining ground truth 6DOF poses for query images is to use SfM. 
An SfM model containing both the database and query images is built and the resulting poses of the query images are used as ground truth~\cite{Li10ECCV,Sun2017CVPR,Sattler2017CVPR}. 
Yet, this approach again relies on local feature matches and can only succeed if the query and database images are sufficiently similar~\cite{Radenovic2016CVPR}. 
The benchmark datasets constructed this way thus tend to only include images that are relatively easy to localize in the first place.

In this paper, we construct the first datasets for benchmarking visual localization under changing conditions. 
To overcome the above mentioned problem, we heavily rely on human work: 
We manually annotate matches between images captured under different conditions and verify the resulting ground truth poses. 
We create three complimentary benchmark datasets   based on existing data~\cite{Sattler12BMVC,Maddern2017IJRR,Badino_IV11}. All consist of a 3D model constructed under one condition and offer query images taken under different conditions: 
The \emph{Aachen Day-Night} dataset focuses on localizing high-quality night-time images against a day-time 3D model. 
The \emph{RobotCar Seasons} and \emph{CMU Seasons} dataset both consider automotive scenarios and depict the same scene under varying seasonal and weather conditions. 
One challenge of the RobotCar Seasons dataset is to localize low-quality night-time images. 
The CMU Seasons dataset focuses on the impact of seasons on vegetation and thus the impact of scene geometry changes on localization.

This paper makes the following \textbf{contributions}:
\emph{(i)} We create a new outdoor benchmark complete with ground truth and metrics for evaluating 6DOF visual localization under changing conditions such as illumination (day/night), weather (sunny/rain/snow), and seasons (summer/winter).
Our benchmark covers multiple scenarios, such as pedestrian and vehicle localization, and localization from single and multiple images as well as sequences. 
\emph{(ii)}~We provide an extensive experimental evaluation of state-of-the-art algorithms from both the computer vision and robotics communities on our datasets. 
We show that existing algorithms, including SfM, have severe problems dealing with both day-night changes and seasonal changes in vegetated environments. 
\emph{(iii)} We show the value of querying with multiple images, rather than with individual photos, especially under challenging conditions. 
\emph{(iv)} We make our benchmarks 
publicly available at \url{visuallocalization.net} to stimulate research on long-term visual localization. 

\begin{table*}[t]
\begin{center}
\setlength{\tabcolsep}{4pt}
\scriptsize{
\begin{tabular}{|l|c|c|c|c|c|c|c|c|c|}
\hline
   &  & Image & 3D SfM Model & \multicolumn{2}{c|}{\# Images} & \multicolumn{3}{c|}{Condition Changes} & 6DOF query\\\cline{5-9}
  Dataset & Setting & Capture & (\# Sub-Models) & Database & Query & Weather & Seasons & Day-Night &  poses\\
  \hline
  Alderley Day/Night~\cite{milford2012seqslam} & Suburban & Trajectory & & 14,607 & 16,960 & \checkmark & & \checkmark & \\
  Nordland~\cite{sunderhauf2013we} & Outdoors & Trajectory & &  \multicolumn{2}{c|}{143k} & & \checkmark & &\\
  Pittsburgh~\cite{Torii15PAMI} & Urban & Trajectory & & 254k & 24k & & & &\\
  SPED~\cite{Chen2017ICRA} & Outdoors & Static Webcams & & 1.27M & 120k & \checkmark & \checkmark & \checkmark & \\
  Tokyo 24/7~\cite{Torii2015CVPR} & Urban & Free Viewpoint & & 75,984 & 315 & & & \checkmark & \\
  \hline 
  7 Scenes~\cite{Shotton2013CVPR} & Indoor & Free Viewpoint &  & 26,000 & 17,000 & & & & \checkmark\\
  Aachen~\cite{Sattler12BMVC} & Historic City & Free Viewpoint & 1.54M / 7.28M (1) & 3,047 & 369 & & & & \\
  Cambridge~\cite{Kendall2015ICCV} & Historic City & Free Viewpoint & 1.89M / 17.68M (5) & 6,848 & 4,081 & & & & \checkmark (SfM)\\
  Dubrovnik~\cite{Li10ECCV} & Historic City & Free Viewpoint & 1.89M / 9.61M (1) & 6,044 & 800 & & & & \checkmark (SfM)\\
  Landmarks~\cite{Li12ECCV} & Landmarks & Free Viewpoint & 38.19M / 177.82M (1k) & 204,626 & 10,000 & & & & \\
  Mall~\cite{Sun2017CVPR} & Indoor & Free Viewpoint &  & 682 & 2296 & & & & \checkmark\\
  NCLT~\cite{CarlevarisBianco2016IJRR} & Outdoors \& Indoors & Trajectory &  & \multicolumn{2}{c|}{about 3.8M} & & \checkmark & & \checkmark \\
  Rome~\cite{Li10ECCV} & Landmarks & Free Viewpoint & 4.07M / 21.52M (69) & 15,179 & 1000 & & & & \\
  San Francisco~\cite{Chen2011CVPR,Li12ECCV,Sattler2017CVPR} & Urban & Free Viewpoint & 30M / 149M (1) & 610,773 & 442 & & & & \checkmark (SfM)\\
  Vienna~\cite{Irschara09CVPR} & Landmarks & Free Viewpoint & 1.12M / 4.85M (3) & 1,324 & 266 & & & & \\
  \hline
  \textbf{Aachen Day-Night (ours)} & Historic City & Free Viewpoint & 1.65M / 10.55M (1) & 4,328 & 922 & & & \checkmark & \checkmark\\
  \textbf{RobotCar Seasons (ours)} & Urban & Trajectory & 6.77M / 36.15M (49) & 20,862 & 11,934 & \checkmark & \checkmark & \checkmark & \checkmark \\
  \textbf{CMU Seasons (ours)} & Suburban & Trajectory & 1.61M / 6.50M (17) & 7,159 & 75,335  & \checkmark & \checkmark &   & \checkmark \\
\hline
\end{tabular}
\vspace{-6pt}
}
\end{center}
\caption{Comparison with existing benchmarks for place recognition and visual localization. "Condition Changes" indicates that the viewing conditions of the query images and database images differ. 
For some datasets, images were captured from similar camera trajectories. 
If SfM 3D models are available,  we report the number of sparse 3D points and the number of associated features.
Only our datasets provide a diverse set of changing conditions, reference 3D models, and most importantly ground truth 6DOF poses for the query images.}
\label{tab:benchmarks}%
\end{table*}

\vspace{-3pt}
\section{Related Work}
\label{sec:related_work}
\vspace{-6pt}
\PAR{Localization benchmarks.}
Tab.~\ref{tab:benchmarks} compares our benchmark datasets with existing datasets for both visual localization and  place recognition.  
Datasets for place recognition~\cite{milford2012seqslam,sunderhauf2013we,Torii-PAMI2015,Chen2017ICRA,Torii2015CVPR} often provide query images captured under different conditions compared to the database images. 
However, they neither provide 3D models nor 6DOF ground truth poses. 
Thus, they cannot be used to analyze the impact of changing conditions on pose estimation accuracy. 
In contrast, datasets for visual localization~\cite{Sattler12BMVC,Shotton2013CVPR,Kendall2015ICCV,Li10ECCV,Li12ECCV,Sun2017CVPR,Chen2011CVPR,Sattler2017CVPR,Irschara09CVPR} often provide ground truth poses. 
However, they do not exhibit strong changes between query and database images due to relying on feature matching for ground truth generation. 
A notable exception is the Michigan North Campus Long-Term (NCLT) dataset~\cite{CarlevarisBianco2016IJRR}, providing images captured over long period of time and ground truth obtained via GPS and LIDAR-based SLAM. Yet, it does not cover all viewing conditions captured in our datasets, \eg, it does not contain any images taken at night or during rain. 
To the best of our knowledge, ours are the first datasets providing both a wide range of changing conditions and accurate 6DOF ground truth. 
Thus, ours is the first benchmark that measures the impact of changing conditions on pose estimation accuracy.

Datasets such as KITTI~\cite{geiger2013vision}, TorontoCity~\cite{Wang2017ICCV}, or the M\'{a}laga Urban dataset~\cite{blanco2014malaga} also provide street-level image sequences. 
Yet, they are less suitable for visual localization as only few places are visited multiple times. 

\PAR{3D structure-based localization}
methods~\cite{Li10ECCV,Li12ECCV,Sattler17PAMI,Svarm17PAMI,Zeisl15ICCV,Sattler-ICCV-2015,Liu2017ICCV} establish correspondences between 2D features in a query image and 3D points in a SfM point cloud via descriptor matching.  
These 2D-3D matches are then used to estimate the query's camera pose. 
Descriptor matching 
can be accelerated by prioritization 
~\cite{Li10ECCV,Choudhary12ECCV,Sattler17PAMI} and efficient search algorithms~\cite{Lynen15RSS,Donoser14CVPR}. 
In large or complex scenes, descriptor matches become ambiguous due to locally similar structures found in different parts of the scene~\cite{Li12ECCV}. 
This results in high outlier ratios of up to 99\%, which can be handled by exploiting co-visibility information~\cite{Li12ECCV,Sattler-ICCV-2015,Liu2017ICCV} or via geometric outlier filtering~\cite{Svarm17PAMI,Zeisl15ICCV,Camposeco2017CVPR}. 

We evaluate \emph{Active Search}~\cite{Sattler17PAMI} and the \emph{City-Scale Localization} approach~\cite{Svarm17PAMI}, a deterministic geometric outlier filter based on a known gravity direction, as  representatives for efficient respectively scalable localization methods.

\PAR{2D image-based localization}
%
methods approximate the pose of a query image using the pose of the most similar photo retrieved from an image database. They are often used 
for place recognition \cite{Torii2015CVPR,Arandjelovic16,Sattler-CVPR16,Suenderhauf2015RSS,Chen2017ICRA,lowry2016visual} and loop-closure detection \cite{Cummins08IJRR,galvez2012bags,MurArtal2015TR}.
They remain effective at scale~\cite{Arandjelovic14a,Torii-PAMI2015,Sattler-CVPR16,Sattler2017CVPR} and can be robust to changing conditions \cite{Arandjelovic16,Torii2015CVPR,Suenderhauf2015RSS,Chen2017ICRA,Naseer2017ICRA,Sattler2017CVPR}. 
We evaluate two compact VLAD-based~\cite{Jegou10}  image-level representations: 
DenseVLAD~\cite{Torii2015CVPR} aggregates densely extracted SIFT descriptors~\cite{Arandjelovic12,Lowe04IJCV} 
while NetVLAD~\cite{Arandjelovic16} uses learned features. 
Both are robust against day-night changes~\cite{Torii2015CVPR,Arandjelovic16} and work well at large-scale~\cite{Sattler2017CVPR}.

We also evaluate the de-facto standard approach for loop-closure detection in robotics \cite{engel2014lsd,linegar2015work}, where robustness to changing conditions is critical for long-term autonomous navigation~\cite{Chen2017ICRA,Naseer2017ICRA,milford2012seqslam,Suenderhauf2015RSS,Torii2015CVPR,Linegar2016ICRA}: 
FAB-MAP \cite{Cummins08IJRR} is an image retrieval approach based on the Bag-of-Words (BoW) paradigm~\cite{Sivic03} that explicitly models the co-occurrence probability of different visual words. 

\PAR{Sequence-based}
approaches for image retrieval are used for  loop-closure detection in robotics~
\cite{maddern2012cat, milford2012seqslam, naseer2014robust}. Requiring a matched sequence of images in the correct order significantly reduces false positive rates compared to single-image retrieval approaches, producing impressive results including direct day-night matches with SeqSLAM~\cite{milford2012seqslam}. We evaluate 
OpenSeqSLAM~\cite{sunderhauf2013we} on our benchmark.

Multiple cameras with known relative poses can be modelled as a generalized camera~\cite{Pless2003CVPR}, \ie, a camera with multiple centers of projections. 
Approaches for absolute pose estimation for both multi-camera systems~\cite{Lee2015IJRR} and camera trajectories~\cite{Camposeco2016ECCV} from 2D-3D matches exist. 
Yet, they have never been applied for localization in changing conditions. In this paper, we show 
that using multiple images can significantly improve performance in challenging scenarios.

\PAR{Learning-based localization}
methods have been proposed to solve both loop-closure detection~\cite{sunderhauf2015performance, Suenderhauf2015RSS, milford2015sequence, Chen2017ICRA} and pose estimation~\cite{Kendall2015ICCV,Walch2017ICCV,Clark2017CVPR,Schoenberger2018CVPR}. 
They learn features with stable appearance over time~\cite{Chen2017ICRA,muhlfellner2015summary, Naseer2017ICRA}, train  classifiers for place recognition~\cite{Cao13,Gronat-IJCV16,Linegar2016ICRA,Weyand-ECCV16}, and train CNNs 
to regress 2D-3D matches~\cite{Brachmann2016CVPR,Brachmann2017CVPR,Shotton2013CVPR} or camera poses~\cite{Kendall2015ICCV,Walch2017ICCV,Clark2017CVPR}.

\vspace{-3pt}
\section{Benchmark Datasets for 6DOF Localization}
\label{sec:benchmarks}
\vspace{-2pt}
\noindent
This section describes the creation of our three new benchmark datasets. 
Each dataset is constructed from publicly available data, allowing our benchmarks to cover multiple geographic locations. 
We add ground truth poses for all query images and build reference 3D models (\cf~Fig.~\ref{fig:all_models}) from images captured under a single condition. 

All three datasets present different challenges. 
The \emph{Aachen Day-Night} dataset 
 focuses on localizing night-time photos against a 3D model built from day-time imagery. 
The night-time images, taken with a mobile phone using software HDR post-processing, are of high quality. 
The dataset represents a scenario where images are taken with hand-held cameras, \eg, an augmented reality application. 

\begin{figure}[t]
  \centering
  \includegraphics[width=1\linewidth]{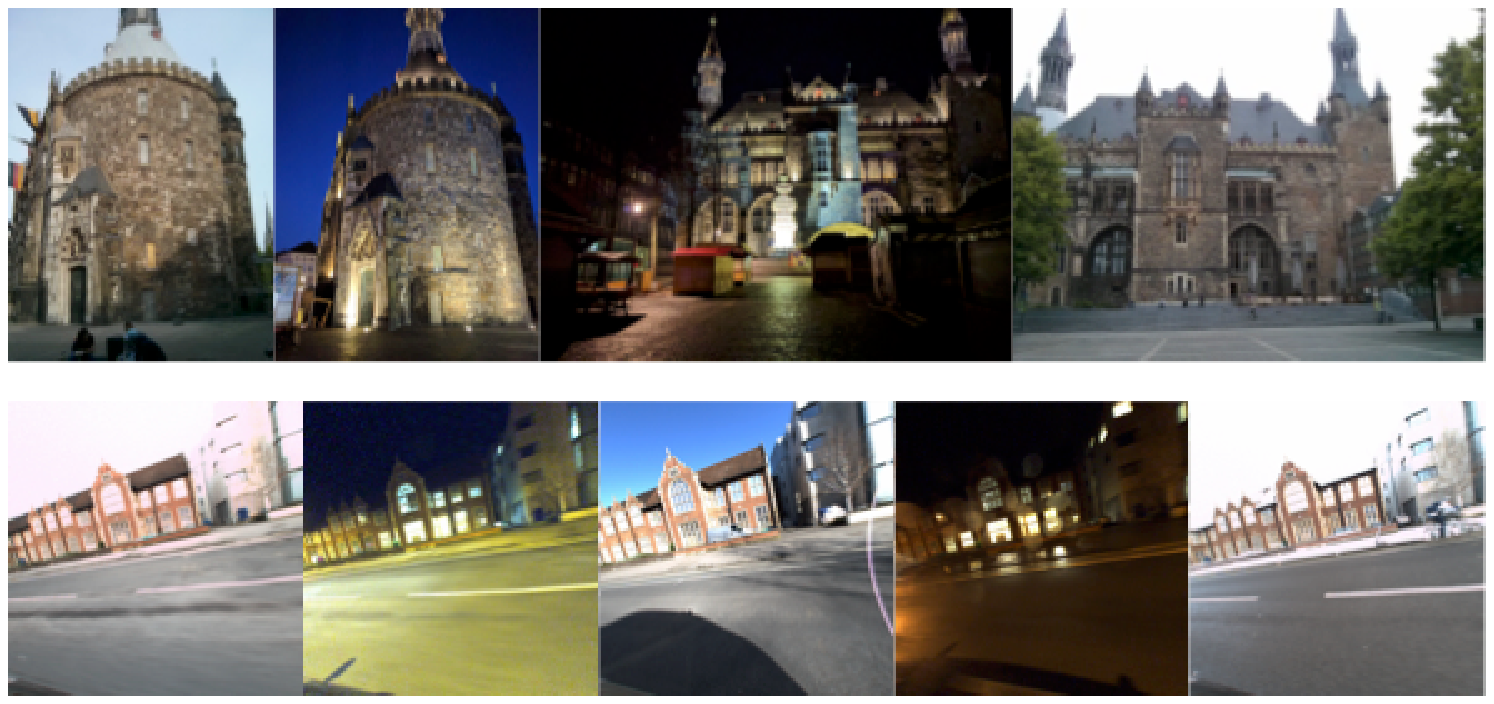}%
  \\
    \includegraphics[width=1\linewidth,clip]{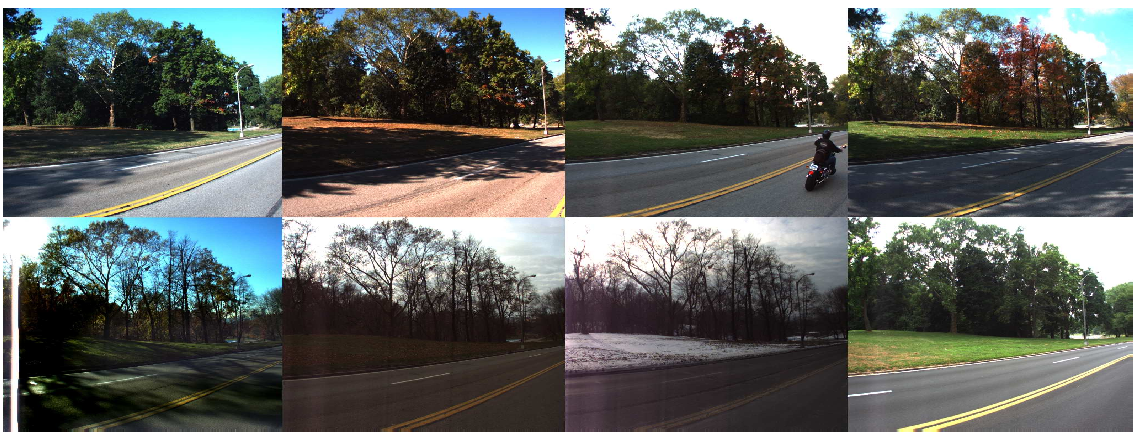}%
\caption{Example query images for \emph{Aachen Day-Night} (top), \emph{RobotCar Seasons} (middle) and \emph{CMU Seasons} datasets (bottom). }
  \label{fig:queries}
\end{figure}

\begin{figure*}[t!]
  \centering
  \includegraphics[width=0.25\linewidth]{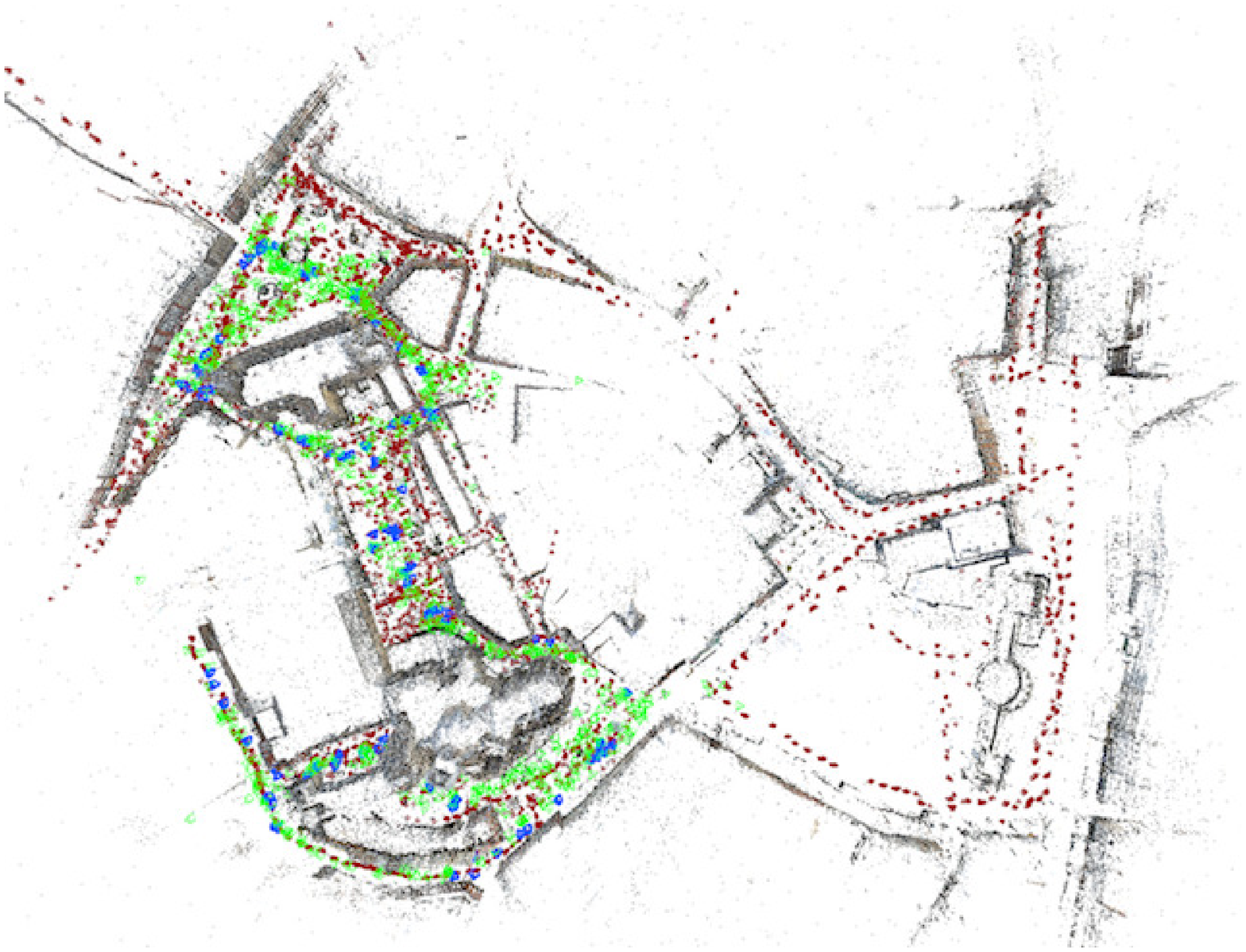}\hspace{32pt}%
  \includegraphics[width=0.20\linewidth]{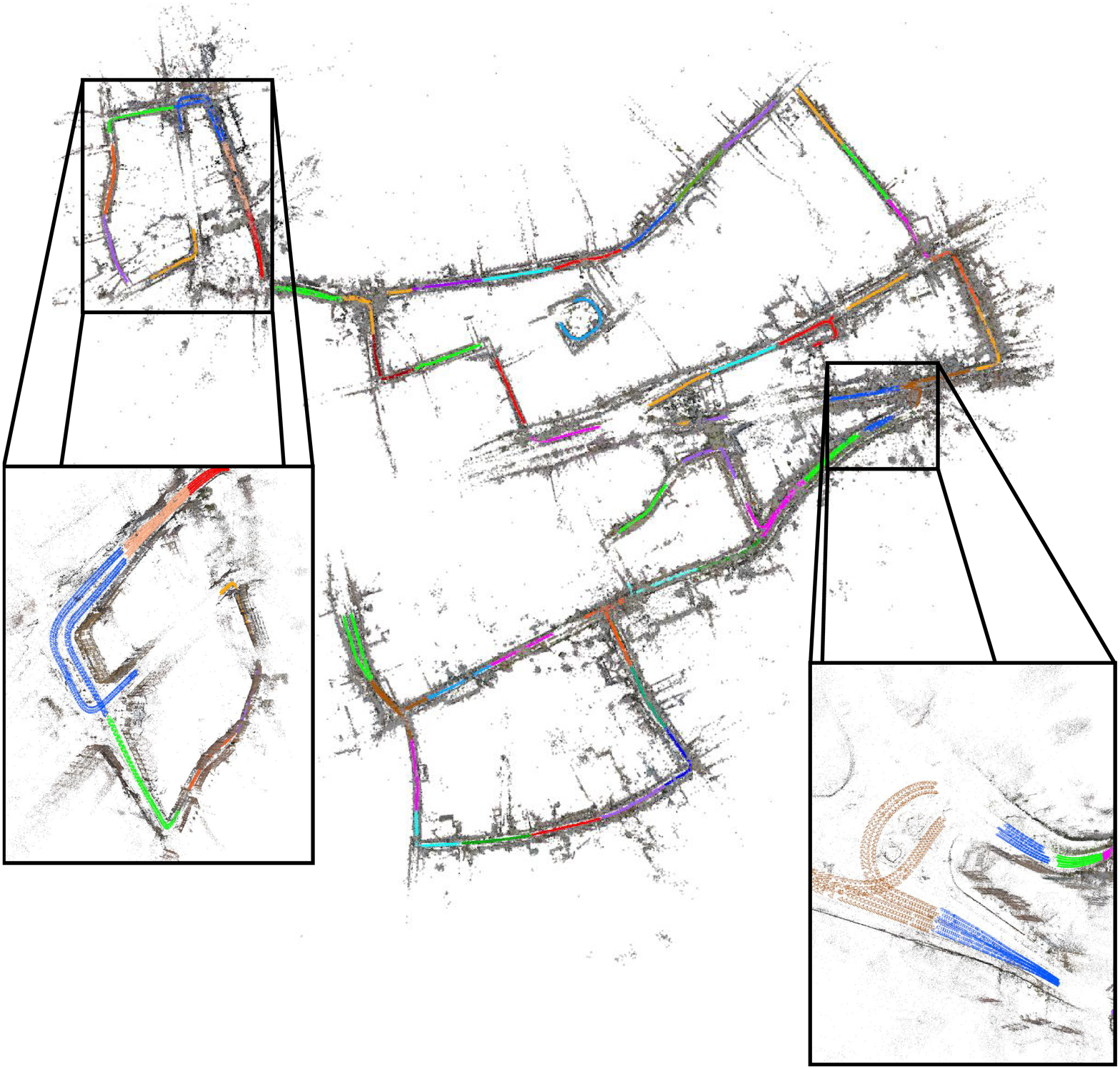}\hspace{32pt}%
  \includegraphics[width=0.23\linewidth]{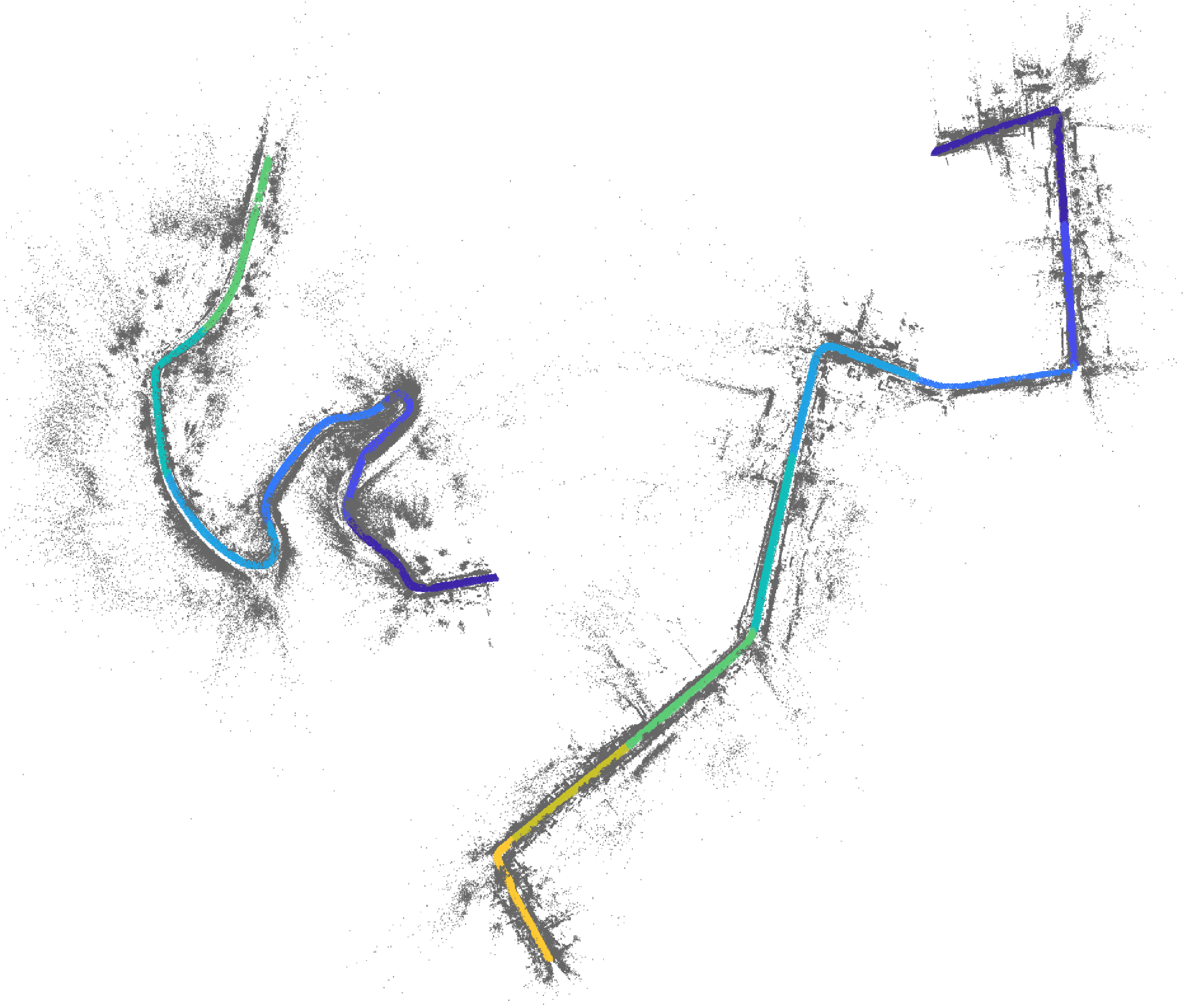}%
  \caption{3D models of the \emph{Aachen Day-Night} (left, showing database (red), day-time query (green), and night-time query images (blue)),  \emph{RobotCar Seasons} (middle), and \emph{CMU Seasons} (right) datasets. 
  For RobotCar and CMU, the colors encode the individual submaps.}
  \label{fig:all_models}
\end{figure*}

Both the \emph{RobotCar Seasons}  
and the \emph{CMU Seasons} datasets 
represent 
automotive scenarios, with images captured from a car. 
In contrast to the Aachen Day dataset, both datasets exhibit less variability in viewpoints but 
a larger variance in viewing conditions. 
The night-time images from the RobotCar dataset were taken from a driving car with a consumer camera with auto-exposure. 
This results in significantly less well-lit images exhibiting motion blur, \ie, images that are significantly harder to localize (\cf Fig.~\ref{fig:queries}). 

The RobotCar dataset depicts a mostly urban scene with rather static scene geometry. 
In contrast, the CMU dataset contains a significant amount of vegetation. 
The changing appearance and geometry of the vegetation, due to seasonal changes, is the main challenge of this dataset.

\begin{table*}[th]
\begin{center}
\scriptsize{
\begin{tabular}{c|c|c|c|c|c}
 & \multicolumn{4}{c|}{reference model} & query images\\ 
 & \# images & \# 3D points & \# features & condition & conditions (\# images) \\
\hline\hline
Aachen Day-Night & 4,328 & 1.65M & 10.55M & day & day (824), night (98)\\ 
\hline
RobotCar Seasons & 20,862 & 6.77M & 36.15M & overcast & dawn (1,449), dusk (1,182), night (1,314), night+rain (1,320), rain (1,263), \\
& & & & (November) & overcast summer / winter (1,389 / 1,170), snow (1,467), sun (1,380)\\
\hline
CMU Seasons & 7,159 & 1.61M & 6.50M & sun / no foliage & sun (22,073), low sun (28,045), overcast (11,383), clouds (14,481), \\
& & & & (April) & foliage (33,897), mixed foliage (27,637), no foliage (13,801) \\
& & & & & urban (31,250), suburban (13,736), park (30,349) \\
\hline
\end{tabular}
\vspace{-6pt}
}
\end{center}
\caption{Detailed statistics for the three benchmark datasets proposed in this paper. For each dataset, a reference 3D model was constructed using images taken under the same reference condition, \eg, "overcast" for the RobotCar Seasons dataset.}%
\label{tab:datasets}%
\end{table*}

\vspace{-3pt}
\subsection{The Aachen Day-Night Dataset}
\label{sec:benchmarks:aachen}
\vspace{-3pt}
\noindent
Our {Aachen Day-Night} dataset is based on the Aachen localization dataset from~\cite{Sattler12BMVC}. 
The original dataset contains 4,479 reference and 369 query images taken 
in the old inner city of Aachen, Germany. 
It provides a 3D SfM model but does not have ground truth poses for the queries. 
We augmented the original dataset with day- and night-time queries captured using standard consumer phone cameras.

To obtain ground truth poses for the day-time queries, we used COLMAP~\cite{Schonberger-CVPR16} to create an intermediate 3D model from the reference and day-time query images. 
The scale of the reconstruction is recovered by aligning it with the geo-registered original Aachen model. 
As in~\cite{Li10ECCV}, we obtain the reference model for the Aachen Day-Night dataset by removing  the day-time query images. 3D points visible in only a single remaining camera were removed as well~\cite{Li10ECCV}. 
The resulting 3D model has 4,328 reference images and 1.65M 3D points triangulated from 10.55M features.

\PAR{Ground truth for night-time queries.} 
We captured 98 night-time query images using a Google Nexus5X phone with software HDR enabled. 
Attempts to include them in the intermediate model resulted in highly inaccurate camera poses due to a lack of sufficient feature matches. 
To obtain ground truth poses for the night-time queries, we thus hand-labelled 2D-3D matches. 
We manually selected a day-time query image taken from a similar viewpoint for each night-time query. 
For each selected day-time query, we projected its visible 3D points from the intermediate model into it. 
Given these projections as reference, we  manually labelled 10 to 30 corresponding pixel positions  in the night-time query. 
Using the resulting 2D-3D matches and the known intrinsics of the camera, 
we estimate the camera poses using a 3-point solver \cite{Fischler81CACM,Kneip11CVPR} 
and non-linear pose refinement. 

To estimate the accuracy for these poses, we measure the mean reprojection error of our hand-labelled 2D-3D correspondences (4.33 pixels for 1600x1200 pixel images) and the pose uncertainty. 
For the latter, we compute multiple poses from a subset of the matches for each image and measure the difference in these poses to our ground truth poses. 
The mean median position and orientation errors are 36cm and 1$^\circ$. 
The absolute pose accuracy that can be achieved by  minimizing a reprojection error depends on the distance of the camera to the scene. 
Given that the images were typically taken 15 or more meters from the scene, we consider the ground truth poses to be reasonably accurate.

\vspace{-3pt}
\subsection{The RobotCar Seasons Dataset}
\label{sec:benchmarks:robotcar}
\vspace{-3pt}
\noindent
Our {RobotCar Seasons} dataset is based on a subset of the publicly available Oxford RobotCar Dataset \cite{Maddern2017IJRR}. 
The original dataset contains over 20M images recorded from an autonomous vehicle platform over 12 months in Oxford, UK.
Out of the 100 available traversals of the 10km route, we select one reference traversal in overcast conditions and nine query traversals that 
 cover a wide range of conditions (\cf Tab.~\ref{tab:datasets}). 
All selected images were taken with the three synchronized global shutter Point Grey Grasshopper2 cameras mounted to the left, rear, and right of the car. 
Both the intrinsics of the cameras and their relative poses are known.

The reference traversal contains 26,121 images taken at 8,707 positions, with 1m between successive positions.
Building a single consistent 3D model from this data is very challenging, both due to sheer size and the lack of visual overlap between the three cameras. 
We thus built 49 non-overlapping local submaps, each covering a 100m trajectory. 
For each submap, we initialized the database camera poses using vehicle positions reported by the inertial navigation system (INS) mounted on the RobotCar.
We then iteratively triangulated 3D points, 
merged tracks, and refined both structure and poses using bundle adjustment. The scale of the reconstructions was recovered by registering them against the INS poses. The reference model contains all submaps and consists of 20,862 reference images and 6.77M 3D points triangulated from 36.15M features. 

We obtained query images by selecting reference positions inside the 49 submaps and gathering all images from the nine query traversals with INS poses within 10m of one of the positions.
This resulted in 11,934 images in total, where triplets of images were captured at 3,978 distinct locations.
We also grouped the queries into 460 temporal sequences based on the timestamps of the images.

\PAR{Ground truth poses for the queries.}
Due to GPS drift, the INS poses cannot be directly used as ground truth. 
Again, there are not enough feature matches between day- and night-time images for SfM. We thus used the LIDAR scanners mounted to the vehicle to build local 3D point clouds for each of the 49 submaps under each condition. These models were then aligned to the LIDAR point clouds of the reference trajectory using ICP~\cite{besl1992method}. 
Many alignments needed to be manually adjusted 
to account for changes in scene structure over time  (often due to building construction and road layout changes). The final median RMS errors between aligned point clouds was under 0.10m in translation and 0.5$^\circ$ in rotation across all locations.
The alignments provided ground truth poses for the query images. 

\vspace{-3pt}
\subsection{The CMU Seasons Dataset}
\label{sec:benchmarks:cmu}
\vspace{-3pt}
\noindent
The {CMU Seasons} Dataset is based on a subset of the CMU Visual Localization Dataset~\cite{Badino_IV11}, which contains more than 100K images recorded by the Computer Vision Group at Carnegie Mellon University  over a period of 12 months in Pittsburgh, PA, USA. 
The images were collected using a rig of two cameras mounted at 45 degree forward/left and forward/right angles on the roof of an SUV. The vehicle traversed an 8.5 km long route through central and suburban Pittsburgh 16 times with a spacing in time of between 2 weeks up to 2 months. 
Out of the 16 traversals, we selected the one from April 4 as the reference, and then 11 query traversals were selected such that they cover the range of variations in seasons and weather that the data set contains. 

\PAR{Ground truth poses for the queries.} As with the RobotCar dataset, the GPS is not accurate enough and the CMU dataset is also too large to build one 3D model from all the images.
The full sequences were split up into 17 shorter sequences, each containing about 250 consecutive vehicle poses. 
For each short sequence, a 3D model was built using bundle adjustment of SIFT points tracked over several image frames. 
The resulting submaps of the reference route were merged with the corresponding submaps from the other traversals by using global bundle adjustment and manually annotated image correspondences. Reprojection errors are within a few pixels for all 3D points and the distances between estimated camera positions and expected ones (based on neighbouring cameras) are under 0.10m. 
The resulting reference model consists of 1.61M 3D points triangulated from 6.50M features in 7,159 database images. We provide 75,335 query images and 187 query sequences. 

\vspace{-3pt}
\section{Benchmark Setup}
\vspace{-3pt}
\noindent
We evaluate state-of-the-art localization approaches on our new benchmark datasets to measure the impact of changing conditions on camera pose estimation accuracy and to understand  how hard robust long-term localization is.

\PAR{Evaluation measures.}  We measure the \emph{pose accuracy} of a method by the deviation 
between the estimated and the ground truth pose. 
The \emph{position error} is measured as the Euclidean distance $\|c_\text{est} - c_\text{gt}\|_2$ between the estimated $c_\text{est}$ and the ground truth position $c_\text{gt}$. 
The absolute \emph{orientation error} $|\alpha|$, measured as an angle in degrees, is computed from the estimated and ground truth camera rotation matrices $\mathtt{R}_\text{est}$ and $\mathtt{R}_\text{gt}$. 
We follow standard practice~\cite{Hartley2013IJCV} and compute 
$|\alpha|$ as $2\cos(|\alpha|)= \text{trace}(\mathtt{R}_\text{gt}^{-1}\mathtt{R}_\text{est}) - 1$, \ie,  
 we measure the minimum rotation angle required to align both rotations~\cite{Hartley2013IJCV}. 

We measure the percentage of query images localized within $X$m and $Y^\circ$ of their ground truth pose. 
We define three pose accuracy intervals by varying the thresholds: 
\emph{High-precision} (0.25m, 2$^\circ$), \emph{medium-precision} (0.5m, 5$^\circ$), and \emph{coarse-precision} (5m, 10$^\circ$). 
These thresholds were chosen to reflect the high accuracy required for autonomous driving. 
We use the intervals (0.5m, 2$^\circ$), (1m, 5$^\circ$), (5m, 10$^\circ$) for the Aachen night-time queries to account for the higher uncertainty in our ground truth poses.
Still, all regimes are more accurate than consumer-grade GPS systems.

\PAR{Evaluated algorithms.}
As discussed in Sec.~\ref{sec:related_work}, we evaluate a set of state-of-the-art algorithms covering the most common types of localization approaches: 
From the class of 3D structure-based methods, we use \emph{Active Search} (AS)~\cite{Sattler2017CVPR} and \emph{City-Scale Localization} (CSL)~\cite{Svarm17PAMI}. 
From the class of 2D image retrieval-based approaches, we use
\emph{DenseVLAD}~\cite{Torii2015CVPR},   \emph{NetVLAD}~\cite{Arandjelovic16}, and \emph{FAB-MAP}~\cite{Cummins08IJRR}.

In order to measure the benefit of using multiple images for pose estimation, we evaluate two approaches: 
\emph{OpenSeqSLAM}~\cite{sunderhauf2013we} is based on image retrieval and enforces that the images in the sequence are matched in correct order. 
Knowing the relative poses between the query images, we can model them as a generalized camera \cite{Pless2003CVPR}. 
Given 2D-3D matches per individual image (estimated via Active Search), we estimate the pose via a generalized absolute camera pose solver~\cite{Lee2015IJRR} inside a RANSAC loop. 
We denote this approach as \emph{Active Search+GC} (AS+GC). 
We mostly use ground truth query poses to compute the relative poses that define the generalized cameras\footnote{Note that Active Search+GC only uses the relative poses between the query images to define the geometry of a generalized camera. It does \emph{not} use any information about the absolute poses of the query images.}. 
Thus, AS+GC provides an upper bound on the number of images that can be localized when querying with  generalized cameras.

The methods discussed above all perform localization from scratch without any prior knowledge about the pose of the query. 
In order to measure how hard our datasets are, we also implemented two \emph{optimistic baselines}. 
Both assume that a set of relevant database images is known for each query. 
Both perform pairwise image matching and use the known ground truth poses for the reference images to triangulate the scene structure. 
The feature matches between the query and reference images and the known intrinsic calibration are then be used to estimate the query pose. 
The first optimistic baseline, \emph{LocalSfM}, uses upright RootSIFT features~\cite{Lowe04IJCV,Arandjelovic12}. 
The second uses upright CNN features densely extracted on a regular grid. We use the same VGG-16 network~\cite{Simonyan15} as NetVLAD. 
The \emph{DenseSfM} method uses coarse-to-fine matching  with conv4 and conv3 features. 

We select the relevant reference images for the two baselines as follows: 
For Aachen, we use the manually selected day-time image (\cf Sec.~\ref{sec:benchmarks:aachen}) to select up to 20 reference images sharing the most 3D points with the selected day-time photo. 
For RobotCar and CMU, we use all reference images within 5m and 135$^\circ$ of the ground truth query pose.

We evaluated \emph{PoseNet}~\cite{Kendall2015ICCV} but were not able to obtain competitive results. We also attempted to train \emph{DSAC}~\cite{Brachmann2017CVPR} on KITTI but were not able to train it. Both PoseNet and DSAC were thus excluded from further evaluations.

\vspace{-3pt}
\section{Experimental Evaluation}
\label{sec:evaluation}
\vspace{-3pt}
\noindent
This section presents the second main contribution of this paper, a detailed experimental evaluation on the effect of changing conditions on the pose estimation accuracy of visual localization techniques. 
In the following, we focus on pose accuracy.
Please see the appendix for experiments concerning computation time. 

\begin{figure}[tbp]
  \centering
  \includegraphics[width=0.98\linewidth]{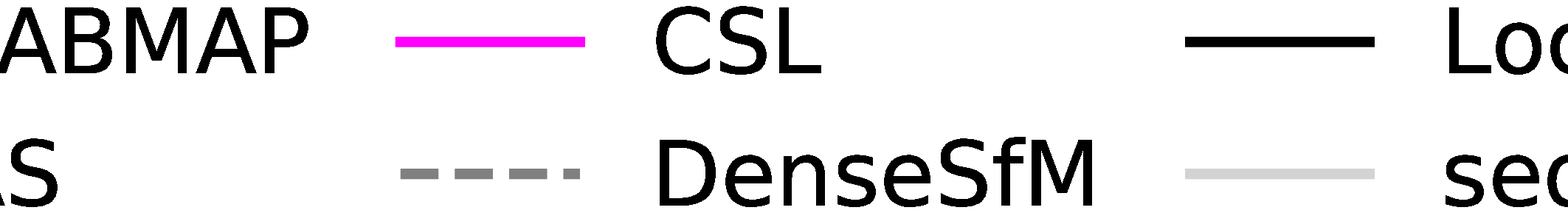}
    \includegraphics[width=0.49\linewidth]{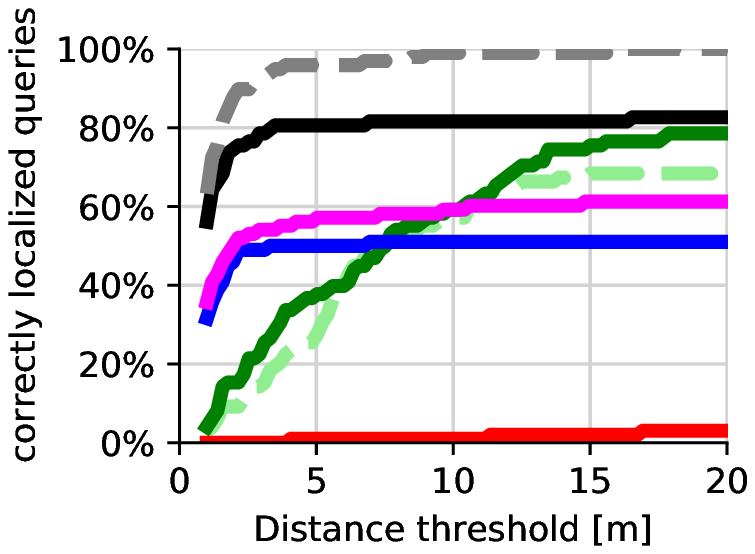} 
    \includegraphics[width=0.49\linewidth]{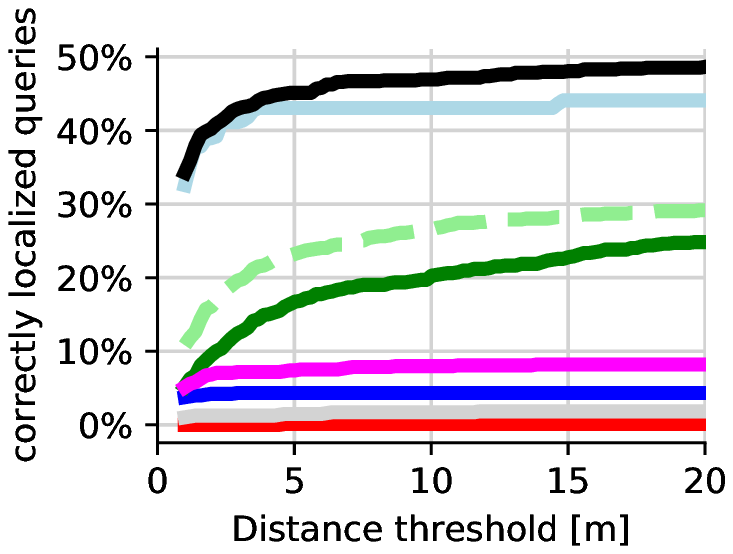} 
  \caption{Cumulative distribution of position errors for the night-time queries of the Aachen (left) and RobotCar (right) datasets.
  }
  \label{fig:results:aachen}
\end{figure}

\begin{table*}[th!]
\begin{center}
\scriptsize{

\setlength\tabcolsep{1.1pt}\begin{tabular}{l|c|c||c|c|c||c|c|c}& \multicolumn{2}{c||}{\textbf{Aachen}}& \multicolumn{6}{c}{\textbf{CMU}}\\
& {day}& {night}& {foliage}& {mixed foliage}& {no foliage}& {urban}& {suburban}& {park}\\ \cline{2-9}
\multicolumn{1}{r|}{\begin{tabular}[c]{@{}r@{}}m\\ deg\end{tabular}} & {\begin{tabular}[c]{@{}c@{}}.25/.50/5.0\\ 2/5/10\end{tabular}} & {\begin{tabular}[c]{@{}c@{}}0.5/1.0/5.0\\ 2/5/10\end{tabular}} & {\begin{tabular}[c]{@{}c@{}}.25/.50/5.0\\ 2/5/10\end{tabular}} & {\begin{tabular}[c]{@{}c@{}}.25/.50/5.0\\ 2/5/10\end{tabular}} & {\begin{tabular}[c]{@{}c@{}}.25/.50/5.0\\ 2/5/10\end{tabular}} & {\begin{tabular}[c]{@{}c@{}}.25/.50/5.0\\ 2/5/10\end{tabular}} & {\begin{tabular}[c]{@{}c@{}}.25/.50/5.0\\ 2/5/10\end{tabular}} & {\begin{tabular}[c]{@{}c@{}}.25/.50/5.0\\ 2/5/10\end{tabular}}\\ \hline
{Active Search} & 57.3 /\083.7 /\096.6 & 19.4 /\030.6 /\043.9 & 28.8 /\0 32.5 /\0 35.9 & 25.1 /\0 29.4 /\0 33.9 & 52.5 /\0 59.4 /\0 66.7 & 55.2 /\0 60.3 /\0 65.1 & 20.7 /\0 25.9 /\0 29.9 & 12.7 /\0 16.3 /\0 20.8\\ \hline
{CSL} & 52.3 /\080.0 /\094.3 & 24.5 /\033.7 /\049.0 & 16.3 /\0 19.1 /\0 26.0 & 15.2 /\0 18.8 /\0 28.6 & 36.5 /\0 43.2 /\0 57.5 & 36.7 /\0 42.0 /\0 53.1 & 8.6 /\0 11.7 /\0 21.1 & 7.0 /\0 9.6 /\0 17.0\\ \hline
{DenseVLAD} & \00.0 /\0\00.1 /\022.8 & \00.0 /\0\02.0 /\014.3 & 13.2 /\0 31.6 /\0 82.3 & 16.2 /\0 38.1 /\0 85.4 & 17.8 /\0 42.1 /\0 91.3 & 22.2 /\0 48.7 /\0 92.8 & 9.9 /\0 26.6 /\0 85.2 & 10.3 /\0 27.0 /\0 77.0\\ \hline
{NetVLAD} & \00.0 /\0\00.2 /\018.9 & \00.0 /\0\02.0 /\012.2 & 10.4 /\0 26.1 /\0 80.1 & 11.0 /\0 26.7 /\0 78.4 & 11.8 /\0 29.1 /\0 82.0 & 17.4 /\0 40.3 /\0 93.2 & 7.7 /\0 21.0 /\0 80.5 & 5.6 /\0 15.7 /\0 65.8\\ \hline
FABMAP & \00.0 /\0\00.0 /\0\04.6 & \00.0 /\0\00.0 /\0\00.0 & 1.1 /\0 2.7 /\0 16.5 & 1.0 /\0 2.5 /\0 14.7 & 3.6 /\0 7.9 /\0 30.7 & 2.7 /\0 6.4 /\0 27.3 & 0.5 /\0 1.5 /\0 13.6 & 0.8 /\0 1.7 /\0 11.5\\ \hline \hline
{LocalSfM} &  & 36.7 /\054.1 /\072.4 & 55.4 /\0 57.0 /\0 59.9 & 52.4 /\0 55.1 /\0 58.6 & 70.8 /\0 72.7 /\0 75.9 & 72.8 /\0 74.1 /\0 76.1 & 55.2 /\0 57.7 /\0 61.3 & 41.8 /\0 44.5 /\0 48.7\\ \hline
{DenseSfM} &  & 39.8 /\060.2 /\084.7 &  &  &  &  &  & \\ \hline
{AS+GC(seq)} &  &  & 86.6 /\0 93.0 /\0 99.3 & 76.3 /\0 88.5 /\0 99.8 & 77.6 /\0 86.8 /\0 99.8 & 86.4 /\0 93.6 /\0 99.8 & 92.0 /\0 96.0 /\0 99.7 & 71.0 /\0 84.0 /\0 99.2\\ \hline
\end{tabular}
}
\end{center}
\vspace{-12pt}
\caption{Evaluation on the \textbf{Aachen Day-Night} dataset and a subset of the conditions of the \textbf{CMU Seasons} dataset.}
\label{tab:results:cmu:comparison}
\end{table*}

\vspace{-3pt}
\subsection{Evaluation on the Aachen Day-Night Dataset}
\vspace{-3pt}
\noindent The focus of the Aachen Day-Night dataset is on benchmarking the pose accuracy obtained by state-of-the-art methods when localizing night-time queries against a 3D model constructed from day-time imagery. 
In order to put the results obtained for the night-time queries into context, we first evaluate a subset of the methods on the 824 day-time queries. 
As shown in Tab.~\ref{tab:results:cmu:comparison}, 
the two structure-based methods are able to estimate accurate camera poses and localize nearly all images within the coarse-precision regime. 
We conclude that the Aachen dataset is not particularly challenging for the day-time query images. 

\PAR{Night-time queries.}
Tab.~\ref{tab:results:cmu:comparison} also reports the results obtained for the night-time queries. 
We observe a significant drop in pose accuracy for both Active Search and CSL, down from above 50\% in the high-precision regime to less than 50\% in the coarse-precision regime. 
Given that the night-time queries were taken from similar viewpoints as the day-time queries, 
this drop 
is solely caused by the day-night change. 

CSL localizes more images compared to Active Search (AS). 
This is not surprising since CSL also uses matches that were rejected by AS as too ambiguous. 
Still, there is a significant difference to LocalSfM. 
CSL and AS both match features against the full 3D model while LocalSfM only considers a small part of the model for each query. 
This shows that global matching sufficiently often fails to find the correct nearest neighbors, likely caused by significant differences between day-time and night-time descriptors. 

Fig.~\ref{fig:results:aachen}(left) shows the cumulative distribution of position errors for the night-time queries and provides interesting insights: 
LocalSfM, despite knowing relevant reference images for each query, completely fails to localize about 20\% of all queries. 
This is caused by a lack of correct feature matches for these queries, either due to failures of the feature detector or descriptor. 
DenseSfM skips feature detection and directly matches densely extracted CNN descriptors (which encode higher-level information compared to the gradient histograms used by RootSIFT). 
This enables DenseSfM to localize more images at a higher accuracy, resulting in the best performance on this dataset. 
Still, there is significant room for improvement, even in the coarse-precision regime (\cf Tab.~\ref{tab:results:cmu:comparison}). 
Also, extracting and matching dense descriptors is a time-consuming task.

\begin{table*}[t!]
\begin{center}
\scriptsize{
\setlength\tabcolsep{1.3pt}
\begin{tabular}{l|c|c|c|c|c|c|c||c|c}
& \multicolumn{7}{c||}{\textbf{day conditions}} & \multicolumn{2}{c}{\textbf{night conditions}}\\ 
& {dawn} & {dusk} & {OC-summer} & {OC-winter} & {rain} & {snow} & {sun} & {night} & {night-rain}\\ \cline{2-10}
\multicolumn{1}{r|}{\begin{tabular}[c]{@{}r@{}}m\\ deg\end{tabular}} & {\begin{tabular}[c]{@{}c@{}}.25 / .50 / 5.0\\ 2 / 5 / 10\end{tabular}} & {\begin{tabular}[c]{@{}c@{}}.25 / .50 / 5.0\\ 2 / 5 / 10\end{tabular}} & {\begin{tabular}[c]{@{}c@{}}.25 / .50 / 5.0\\ 2 / 5 / 10\end{tabular}} & {\begin{tabular}[c]{@{}c@{}}.25 / .50 / 5.0\\ 2 / 5 / 10\end{tabular}} & {\begin{tabular}[c]{@{}c@{}}.25 / .50 / 5.0\\ 2 / 5 / 10\end{tabular}} & {\begin{tabular}[c]{@{}c@{}}.25 / .50 / 5.0\\ 2 / 5 / 10\end{tabular}} & {\begin{tabular}[c]{@{}c@{}}.25 / .50 / 5.0\\ 2 / 5 / 10\end{tabular}} & {\begin{tabular}[c]{@{}c@{}}.25 / .50 / 5.0\\ 2 / 5 / 10\end{tabular}} & {\begin{tabular}[c]{@{}c@{}}.25 / .50 / 5.0\\ 2 / 5 / 10\end{tabular}}\\ \hline
{ActiveSearch} & 36.2 / 68.9 / 89.4 & 44.7 / 74.6 / 95.9 & 24.8 / 63.9 / 95.5 & 33.1 / 71.5 / 93.8 & 51.3 / 79.8 / 96.9 & 36.6 / 72.2 / 93.7 & 25.0 / 46.5 / 69.1 & 0.5 / 1.1 / \03.4 & 1.4 / 3.0 / \05.2\\ \hline
{CSL} & \047.2 /73.3 / 90.1 & 56.6 / 82.7 / 95.9 & 34.1 / 71.1 / 93.5 & 39.5 / 75.9 / 92.3 & 59.6 / 83.1 / 97.6 & 53.2 / 83.6 / 92.4 & 28.0 / 47.0 / 70.4 & 0.2 / 0.9 / \05.3 & 0.9 / 4.3 / \09.1\\ \hline
{DenseVLAD} & \08.7 / 36.9 / 92.5 & 10.2 / 38.8 / 94.2 & \06.0 / 29.8 / 92.0 & \04.1 / 26.9 / 93.3 & 10.2 / 40.6 / 96.9 & \08.6 / 30.1 / 90.2 & \05.7 / 16.3 / 80.2 & 0.9 / 3.4 / 19.9 & 1.1 / 5.5 / 25.5\\ \hline
{NetVLAD} & \06.2 / 22.8 / 82.6 & \07.4 / 29.7 / 92.9 & \06.5 / 29.6 / 95.2 & \02.8 / 26.2 / 92.6 & \09.0 / 35.9 / 96.0 & \07.0 / 25.2 / 91.8 & \05.7 / 16.5 / 86.7 & 0.2 / 1.8 / 15.5 & 0.5 / 2.7 / 16.4\\ \hline
{FABMAP} & \01.2 / \05.6 / 14.9 & \04.1 / 18.3 / 55.1 & \00.9 / \08.9 / 39.3 & \02.6 / 13.3 / 44.1 & \08.8 / 32.1 / 86.5 & \02.0 / \08.2 / 28.4 & \00.0 / \00.0 / \02.4 & 0.0 / 0.0 / \00.0 & 0.0 / 0.0 / \00.0\\ \hline
\end{tabular}

}
\end{center}
\vspace{-6pt}
\caption{Evaluation on the \textbf{RobotCar Seasons} dataset. We report the percentage of queries localized within the three thresholds.} \label{tab:results:robotcar:comparison}
\end{table*}

\vspace{-3pt}
\subsection{Evaluation on the RobotCar Seasons Dataset}
\vspace{-3pt}
\noindent
The focus of the RobotCar Seasons dataset is to measure the impact of different seasons and illumination conditions on pose estimation accuracy in an urban environment.

{Tab.~\ref{tab:results:robotcar:comparison} shows that changing day-time conditions have only a small impact on pose estimation accuracy for all methods.}
The reason is that seasonal changes have little impact on the building facades that are dominant in most query images.
The exceptions are ``dawn'' and ``sun''. For both, we observed overexposed images caused by direct sunlight 
(\cf Fig.~\ref{fig:robotcar-projection-1}). 
Thus, fewer features can be found for Active Search and CSL and the global image descriptors used by the image retrieval approaches are affected as well. 

On the Aachen Day-Night dataset, we observed that image retrieval-based methods (DenseVLAD and NetVLAD) consistently performed worse than structure-based methods (Active Search, CSL, LocalSfM, and DenseSfM). For the RobotCar dataset, NetVLAD and DenseVLAD essentially achieve the same coarse-precision performance as Active Search and CSL. 
This is caused by the lower variation in viewpoints as the car follows the same road. 

Compared to Aachen, 
there is an even stronger drop in pose accuracy between day and night for the RobotCar dataset. All methods fail to localize a significant number of queries for both the high- and medium-precision regimes. 
Interestingly, DenseVLAD and NetVLAD outperform all other methods in the coarse-precision regime (\cf~Fig.~\ref{fig:results:aachen}(right)). 
This shows that their global descriptors still encode distinctive information even if 
local feature matching fails. 
The better performance of all methods under "night+rain" compared to "night" comes from the autoexposure of the RobotCar's cameras. A longer exposure is used for the "night", leading to significant motion blur. 

\begin{table}[t]
\begin{center}
\scriptsize{
\setlength\tabcolsep{3.0pt}\begin{tabular}{l|c|c}
& \textbf{all day}& \textbf{all night}\\ \cline{2-3}
\multicolumn{1}{r|}{\begin{tabular}[c]{@{}r@{}}m\\ deg\end{tabular}} & {\begin{tabular}[c]{@{}c@{}}.25 / .50 / 5.0\\ 2 / 5 / 10\end{tabular}} & {\begin{tabular}[c]{@{}c@{}}.25 / .50 / 5.0\\ 2 / 5 / 10\end{tabular}}\\ \hline
{ActiveSearch} & 35.6 / 67.9 / 90.4 & \00.9 / \02.1 / \04.3 \\ \hline
{CSL} & 45.3 / 73.5 / 90.1 & \00.6 / \02.6 / \07.2 \\ \hline \hline
{ActiveSearch+GC (triplet)} & 45.5 / 77.0 / 94.7 & \02.7 / \06.9 / 12.1 \\ \hline
{ActiveSearch+GC (sequence, GT)} & 46.7 / 80.1 / 97.0 & \05.8 / 21.0 / 43.1 \\ \hline
{seqSLAM} & \01.3 / \06.1 / 15.3 & \00.2 / \00.7 / \01.5 \\ \hline 
\end{tabular}

}
\end{center}
\vspace{-6pt}
\caption{
Using \textbf{multiple images} for pose estimation (ActiveSeach+GC) on the \textbf{RobotCar Seasons} dataset.} 
\label{tab:results:robotcar:multicam}
\end{table}

\PAR{Multi-image queries.} The RobotCar is equipped with three synchronized cameras and captures sequences of images for each camera. Rather than querying with only a single image, we can thus also query with multiple photos. 
Tab.~\ref{tab:results:robotcar:multicam} shows the results obtained with seqSLAM (which uses temporal sequences of all images captured by the three cameras) and Active Search+GC. For the latter, we query with triplets of images taken at the same time as well as with temporal sequences of triplets. 
For the triplets, we use the known extrinsic calibration between the three cameras mounted on the car. 
For the temporal sequences, we use relative poses obtained from the ground truth (GT) absolute poses. 
For readability, we only show the results summarized for day- and night-conditions. 

Tab.~\ref{tab:results:robotcar:multicam} shows that Active Search+GC consistently outperforms single image methods in terms of pose accuracy. 
Active Search+GC is able to accumulate correct matches over multiple images. 
This enables Active Search+GC to succeed even if only a few matches are found for each individual image. 
Naturally, the largest gain can be observed when using multiple images in a sequence. 

\begin{table}[t]
\begin{center}
\scriptsize{
\setlength\tabcolsep{3.0pt}\begin{tabular}{lc|c}
&  & \textbf{RobotCar - all night}\\ \cline{3-3}
\multicolumn{2}{r|}{\begin{tabular}[c]{@{}r@{}}m\\ deg\end{tabular}} & {\begin{tabular}[c]{@{}c@{}}.25 / .50 / 5.0\\ 2 / 5 / 10\end{tabular}}\\ \hline
\multirow{2}{*}{ActiveSearch} & full model & \00.9 / \02.1 / \04.3 \\
 & sub-model & \03.2 / \07.9 / 12.0 \\ \hline
\multirow{2}{*}{CSL} & full model & \00.6 / \02.6 / \07.2 \\
 & sub-model & \00.5 / \02.8 / 13.4\\ \hline
\multirow{2}{*}{ActiveSearch+GC (triplet)} & full model & \02.7 / \06.9 / 12.1 \\
 & sub-model & \07.4 / 15.3 / 27.0 \\ \hline
\multirow{2}{*}{ActiveSearch+GC (sequence, GT)} & full model & \05.8 / 21.0 / 43.1 \\
 & sub-model & 13.3 / 35.9 / 61.8 \\ \hline
\multirow{2}{*}{ActiveSearch+GC (sequence, VO)} & full model & \01.5 / \07.4 / 22.9 \\
 & sub-model & \03.6 / 12.5 / 42.2 \\ \hline 
{LocalSfM} & sub-model & 16.1 / 27.3 / 44.1 \\ \hline
\end{tabular}
}
\end{center}
\vspace{-6pt}
\caption{Using \textbf{location priors} to query only submodels rather than the full \textbf{RobotCar Seasons} dataset for night-time queries.} 
\label{tab:results:robotcar:priors}
\end{table}

\PAR{Location priors.} In all previous experiments, we considered the full RobotCar 3D model for localization. 
However, it is not uncommon in outdoor settings to have a rough prior on the location at which the query image was taken. We simulate such a prior by only considering the sub-model relevant to a query rather than the full model. 
While we observe  only a small improvement for day-time queries, localizing night-time queries significantly benefits from solving an easier matching problem (\cf Tab.~\ref{tab:results:robotcar:priors}). For completeness, we also report results for LocalSfM, which also considers only a small part of the model relevant to a query. 
Active Search+GC outperforms LocalSfM on this easier matching task when querying with sequences. 
This is due to not relying on one single image to provide enough matches. 

One drawback of sequence-based localization is that the relative poses between the images in a sequence need to be known quite accurately. 
Tab.~\ref{tab:results:robotcar:priors} also reports results obtained when using our own multi-camera visual odometry (VO) system to compute the relative poses. 
While performing worse compared to ground truth relative poses, this more realistic baseline still outperforms methods using individual images.
The reasons for the performance drop are drift and collapsing trajectories due to degenerate configurations.

\vspace{-3pt}
\subsection{Evaluation on the CMU Seasons Dataset}
\vspace{-3pt}
\noindent
Compared to the urban scenes shown in the other datasets, significant parts of the CMU Seasons dataset show suburban or park regions. 
Seasonal changes can drastically affect the appearance of such regions. In the following, we thus focus on these conditions (see the appendix for an evaluation of all conditions). 
For each query image, we only consider its relevant sub-model. 

Tab.~\ref{tab:results:cmu:comparison} evaluates the impact of changes in foliage and of different regions on pose accuracy. 
The reference condition for the CMU Seasons dataset does not contain foliage. 
Thus, other conditions for which foliage is also absent lead to the most accurate poses. 
Interestingly, DenseVLAD and NetVLAD achieve a better performance than Active Search and CSL for the medium- and coarse-precision regimes under the "Foliage" and "Mixed Foliage" conditions. For the coarse-precision regime, they even outperform LocalSfM. 
This again shows that global image-level descriptors can capture information lost by local features. 

We observe a significant drop in pose accuracy in both suburban and park regions. 
This is caused by the dominant presence of vegetation, leading to many locally similar (and thus globally confusing) features. 
LocalSfM still performs well 
as it only considers a few reference images that are known to be a relevant for a query image. 
Again, we notice that DenseVLAD and NetVLAD are able to coarsely localize more queries compared to the feature-based methods.

Localizing sequences (Active Search+GC) again drastically helps to improve pose estimation accuracy. 
Compared to the RobotCar Seasons dataset, where the sequences are rather short (about 20m maximum), the sequences used for the CMU Seasons dataset completely cover their corresponding sub-models. 
In practical applications, smaller sequences are preferable to avoid problems caused by drift when estimating the relative poses in a sequence. 
Still, the results from Tab.~\ref{tab:results:cmu:comparison} show the potential of using multiple rather than a single image for camera pose estimation. 

\vspace{-3pt}
\section{Conclusion \& Lessons Learned}
\vspace{-3pt}
\noindent
In this paper, we have introduced three challenging new benchmark datasets for visual localization, allowing us, 
for the first time, to analyze the impact of changing conditions on the accuracy of 6DOF camera pose estimation. Our experiments clearly show that the long-term visual localization problem is far from solved. 

The extensive experiments performed in this paper lead to multiple interesting conclusions: 
(i) Structure-based methods such as Active Search and CSL are robust to most viewing conditions in urban environments. 
Yet, performance in the high-precision regime still needs to be improved significantly. 
(ii) Localizing night-time images against a database built from day-time photos is a very challenging problem, even when a location prior is given. 
(iii)~Scenes with a significant amount of vegetation are challenging, even when a location prior is given. 
(iv) SfM, typically used to obtain ground truth for localization benchmarks, does not fully handle problems (ii) and (iii) due to limitations of existing local features. 
Dense CNN feature matching inside SfM improves pose estimation performance at high computational costs, but does not fully solve the problem. 
Novel (dense) features, \eg, based on scene semantics~\cite{Schoenberger2018CVPR}, seems to be required to solve these problems. 
Our datasets readily provide a benchmark for such features through the LocalSfM and DenseSfM pipelines. 
(v) Image-level descriptors such as DenseVLAD can succeed in scenarios where local feature matching fails. 
They can even provide coarse-level pose estimates in  autonomous driving scenarios. 
Aiming to improve pose accuracy, \eg, by denser view sampling via synthetic images~\cite{Torii2015CVPR} or image-level approaches for relative pose estimation, is an interesting research direction. 
(vi) There is a clear benefit in using multiple images for pose estimation. Yet, there is little existing work on multi-image localization. 
Fully exploiting the availability of multiple images 
(rather than continuing the focus on single images) is thus another promising avenue for future research.

{\small
 \PAR{Acknowledgements.} This work was partially supported by ERC grant LEAP No.\ 336845, CIFAR Learning in Machines $\&$ Brains program, EU-H2020 project LADIO~731970, the European Regional Development Fund under the project IMPACT (reg. no.\ CZ$.02.1.01/0.0/0.0/15\_003/0000468$), JSPS KAKENHI Grant Number 15H05313, EPSRC Programme Grant EP/M019918/1, the Swedish Research Council (grant no.\ 2016-04445), the Swedish Foundation for Strategic Research (Semantic Mapping and Visual Navigation for Smart Robots), and Vinnova / FFI (Perceptron, grant no.\ 2017-01942).
}

\appendix
\section*{Appendix}
\noindent This appendix provides additional results, in particular evaluations under all conditions on the CMU Seasons dataset and run-time results for the evaluated methods. 
In addition, a more detailed description of the state-of-the-art localization approaches evaluated in the paper is provided. This includes details on the parameter settings used in our experiments, which are provided to foster reproducibility. 

The appendix is structured as follows:
Sec.~\ref{sec:sota} provides a more detailed description of all evaluated state-of-the-art approaches. 
Sec.~\ref{sec:datasets} provides additional details for the RobotCar Seasons and CMU Seasons datasets. 
Sec.~\ref{sec:timings} provides timing results for these methods on the different datasets. 
Sec.~\ref{sec:cmu} shows evaluation results on the \emph{CMU Seasons} under all conditions. 
Finally, Sec.~\ref{sec:distributions} shows the cumulative distributions in position and orientation error for all state-of-the-art methods evaluated on our benchmark.

\section{Details on the Evaluated Algorithms}
\label{sec:sota}
\noindent This section provides a detailed description, including parameter settings, of the state-of-the-art algorithms used for experimental evaluation (\cf Sec.~5 in the paper). 

\subsection{3D Structure-based Localization}
\PAR{Active Search (AS).}
Active Search \cite{Sattler17PAMI} accelerates 2D-3D descriptor matching via a prioritization scheme. 
It uses a visual vocabulary to quantize the descriptor space. 
For each query feature, it determines how many 3D point descriptors are assigned to the feature's closest visual word. 
This determines the number of descriptor comparisons needed for matching this feature. 
Active Search then matches the features in ascending order of the number of required descriptor comparisons. 
If a 2D-to-3D match is found, Active Search attempts to find additional 3D-to-2D correspondences for the 3D points surrounding the matching point. 
Correspondence search terminates once 100 matches have been found. 

For the Aachen Day-Night dataset, we trained a visual vocabulary containing 100k words using approximate k-means clustering \cite{Philbin07} on all upright RootSIFT~\cite{Arandjelovic12,Lowe04IJCV} descriptors found in 1,000 randomly selected database images contained in the 3D model. Similarly, we trained a vocabulary containing 10k words for the RobotCar Seasons dataset from the descriptors found in 1,000 randomly selected images contained in the reference 3D model. 
For the CMU Seasons dataset, we also trained a visual vocabulary consisting of 10k words, but used the SIFT~\cite{Lowe04IJCV} features corresponding to the 3D points in all sub-models instead of RootSIFT features. \TPdel{All vocabularies do not contain any information from the query images.} No vocabulary contains any information from the query images.

We use calibrated cameras rather than simultaneously estimating each camera's extrinsic and intrinsic parameters.
We thereby exploit the known intrinsic calibrations provided by the intermediate model of the Aachen Day-Night dataset\footnote{Some of the day-time queries were taken with the same camera as the night-time queries and we enforced that the images taken with the same camera have consistent intrinsics. Thus, the intermediate model provides the intrinsic calibration of the night-time queries.} and the known intrinsics of the RobotCar Seasons and CMU Seasons datasets.

Besides training new vocabularies and using calibrated cameras, we only changed the threshold on the re-projection error used by RANSAC to distinguish between inliers and outliers. For the Aachen Day-Night dataset, we used a threshold of 10 pixels while we used 5 pixels for both the RobotCat Seasons and the CMU Seasons datasets. Otherwise, we used the standard parameters of Active Search.

\PAR{City-Scale Localization (CSL).}
The City-Scale Localization algorithm~\cite{Svarm17PAMI} is an outlier rejection algorithm, \ie, it is a robust localization algorithm that can prune guaranteed outlier correspondences from a given set of 2D-3D correspondences. 
CSL is based on the following central insight: If the gravity direction and an approximate height of the camera above the ground plane are known, it is possible to  calculate an upper bound for the maximum number of inliers that any solution containing a given 2D-3D correspondence as an inlier can have. 
At the same time, CSL also computes a lower bound on the number of inliers for a given correspondence by computing a pose for which this correspondence is an inlier.
CSL thus computes this upper bound for each 2D-3D match and, similar to RANSAC, continuously updates the best pose found so far (which provides a lower bound on the number of inliers that can be found). 
All correspondences with an upper bound on the maximum number of inliers that is below the number of inliers in the current best solution can be permanently discarded from further consideration. When outliers have been discarded, three-point RANSAC~\cite{Fischler81CACM,Kneip11CVPR} is performed on the remaining correspondences. 
Notice that, unlike RANSAC, the outlier filter used by CSL is deterministic. The computational complexity of the filter is $\mathcal{O}(N^2\log N)$, where $N$ is the number of available 2D-3D correspondences.

In order to obtain an estimate for the gravity direction, we follow~\cite{Svarm17PAMI} and add noise to the gravity direction obtained from the ground truth poses. 
CSL iterates over a range of plausible height values, similar to~\cite{Zeisl15ICCV}. In these experiments, the height values cover an interval five meters high. This interval is centered on the camera height of the ground truth pose, with added noise. In the Aachen experiments, the height interval is divided into nine sections, and for the Oxford and CMU experiments, the height interval is divided into three sections. 

The 2D-3D correspondences are generated by matching the descriptors of all detected features in the query image to the descriptors of the 3D points using approximate nearest neighbour search. To account for the fact that each 3D point is associated with multiple descriptor, the 3D points are each assigned a single descriptor vector equal to the mean of all its corresponding descriptors. This matching strategy yields the same number of correspondences as the number of detected features. 

As with Active Search, we use a re-projection error threshold of 10 pixels for the Aachen Day-Night dataset and 5 pixels for both the RobotCat Seasons and the CMU Seasons datasets.

\subsection{2D Image-based Localization}

\begin{table}
\centering
\begin{tabular}{c|c}
\textbf{Parameter} & \textbf{Value} \\ \hline
Feature Type & Dense RootSIFT \\ \hline
Vocabulary Size & 128 \\
(trained on SF) & ~ \\ \hline
Descriptor Dimension & 4,096 \\
(after PCA \& whitening) & ~\\
\end{tabular}
\caption{DenseVLAD parameters.}
\label{tab:densevlad-parameters}
\end{table}

\begin{table}
\centering
\begin{tabular}{c|c}
\textbf{Parameter} & \textbf{Value} \\ \hline
Network model & VGG-16 + NetVLAD \\
(trained on Pitts30k) & + whitening \\ \hline
Descriptor Dimension & 4,096 \\
\end{tabular}
\caption{NetVLAD parameters.}
\label{tab:netvlad-parameters}
\end{table}

\PAR{DenseVLAD and NetVLAD.} We use the original implementations of DenseVLAD~\cite{Torii-CVPR15} and NetVLAD~\cite{Arandjelovic16} provided by the authors. Images were processed at their original resolution unless any dimension exceeded $1920$ pixels. For DenseVLAD, we used the Dense SIFT implementation, followed by RootSIFT normalization~\cite{Arandjelovic12}, available in VLFeat~\cite{Vedaldi10a}. The visual vocabulary used consisted of $128$ visual words (centroids) pre-computed on the San-Francisco (SF) dataset~\cite{Chen11b}, \ie, we used a general vocabulary trained on a different yet similiar dataset. For NetVLAD we used the pre-computed network ``Pitts30k'' trained on the Pittsburgh time-machine street-view image dataset~\cite{Arandjelovic16}. The network is therefore not fine-tuned on our datasets, \ie, we again used a general network trained on a different city. 

Given a DenseVLAD or NetVLAD descriptor, we find the most similar reference image by exhaustive nearest neighbor search. While this stage could be accelerated by approximate search, we found this to be unnecessary as the search for a single query descriptor typically takes less than 20ms.

Tables~\ref{tab:densevlad-parameters} and~\ref{tab:netvlad-parameters} summarize the parameters used for DenseVLAD and NetVLAD in our experiments.

\begin{table}
\centering
\begin{tabular}{c|c}
\textbf{Parameter} & \textbf{Value} \\ \hline
Feature Type & UprightSURF128 \\ \hline
Aachen Vocabulary Size & 3585 \\ \hline
RobotCar Vocabulary Size & 5031 \\ \hline
CMU Vocabulary Size & 4847 \\ \hline
$p\left(z_{i}\left|\right.\bar{e}_{i}\right)$ & 0 \\ \hline
$p\left(\bar{z}_{i}\left|\right.e_{i}\right)$ & 0.61 \\ \hline
$p\left(L_{\textrm{new}}\left|\right.Z^{k-1}\right)$ & 0.9
\end{tabular}
\caption{FAB-MAP parameters.}
\label{tab:fabmap-parameters}
\end{table}

\PAR{FAB-MAP.} For FAB-MAP~\cite{Cummins08IJRR}, we trained a separate vocabulary for each location using Modified Sequential Clustering~\cite{teynor2007fast} on evenly spaced database images, resulting in 3,585 visual words for Aachen Day-Night, 5,031 for RobotCar Seasons and 4,847 for CMU Seasons. A Chow-Liu tree was built for each dataset using the Bag-of-Words generated for each database image using the vocabulary. We used the mean field approximation for the new place likelihood (as additional training images were not available for the sampled approach used in \cite{cummins2011appearance}) and the fast lookup-table implementation in \cite{glover2012openfabmap} to perform image retrieval for each of the query locations. Tab.~\ref{tab:fabmap-parameters} summarizes the parameters used for the experiments.

\subsection{Optimistic Baselines}
\noindent As explained in Sec.~5 of the paper, we implemented two \emph{optimistic baselines}. 
Whereas all other localization algorithms evaluated in the paper use no prior information on a query image's pose, both optimistic baselines are given additional knowledge. 
For each query image, we provide a small set of reference images depicting the same part of the model. 
The remaining problem is to establish sufficiently many correspondences between the query and the selected reference images to facilitate  camera pose estimation. 
Thus, both approaches measure an upper bound on the pose  quality that can be achieved with a given type of local feature.

\PAR{LocalSfM.}
Given a query image and its relevant set of reference images, LocalSfM first extracts upright RootSIFT~\cite{Lowe04IJCV,Arandjelovic12} features. 
Next, LocalSfM performs exhaustive feature matching between the relevant reference images as well as between the query and the relevant reference images. 
While Active Seach and CSL both use Lowe's ratio test\footnote{Active Search uses a ratio test threshold of 0.7 for 2D-to-3D and a threshold of 0.6 for 3D-to-2D matching.}, DenseSfM neither uses the ratio test nor a threshold on the maximum descriptor distance. Instead, it only requires matching features to be mutual nearest neighbors. 
Given the known poses and intrinsics for the reference images, LocalSfM triangulates the 3D structure of the scene using the previously established 2D-2D matches.
Notice that the resulting 3D model is automatically constructed in the global coordinate system of the reference 3D model. 
Finally, we use the known intrinsics of the query image and the feature matches between the query and the reference images to estimate the camera pose of the query. 

For each query image, the relevant set of reference images is selected as follows: 
For the RobotCar Seasons and CMU Seasons datasets, we use the ground truth pose of each query image to identify a relevant set of reference images. More precisely, we select all reference images whose camera centers are within 5m of the ground truth position of the query and whose orientations are within 135$^\circ$ of the orientation of the query image. 


As explained in Sec.~3.2 of the paper, we manually select a day-time query image taken from a similar viewpoint for each nigh-time query photo in the Aachen Day-Night dataset. 
The day-time queries were included when constructing the intermediate model. 
Thus, their ground truth poses as well as a set of 3D points visible in each of them are obtain from the intermediate Structure-from-Motion model. 
For each day-time query, we select up to 20 reference images that observe the largest number of the 3D points visible in the day-time query. 
These reference images then form the set of relevant images for the corresponding night-time query photo. 



LocalSfM is implemented using COLMAP~\cite{Schonberger-CVPR16}. 
It is rather straight-forward to replace upright RootSIFT features with other types of local features. 
In order to encourage the use of our benchmark for the evaluation of local features, we will make our implementation publicly available. 


\PAR{DenseSfM.}
%
 DenseSfM modifies the LocalSfM approach by replacing RootSIFT~\cite{Arandjelovic12} features extracted at DoG extrema~\cite{Lowe04IJCV} with features densely extracted from a regular grid~\cite{Bosch07a,Liu08}.
 The goal of this approach is to increase the robustness of feature matching between day- and night-time images~\cite{Torii-CVPR15,Zhou2016ECCVW}. We used convolutional layers (conv4 and conv3) from a VGG-16 network~\cite{Simonyan15}, which was pre-trained as part of the NetVLAD model (Pitts30k), as features.
 We generated tentative correspondences by matching the extracted features in a coarse-to-fine manner: 
 We first match conv4 features and use the resulting matches to restrict the correspondence search for conv3 features. 
 As for LocalSfM, we performed exhaustive pairwise image matching.
 The matches are verified by estimating up to two homographies between each image pair via RANSAC~\cite{Fischler81CACM}.  The resulting verified feature matches are then used as input for COLMAP~\cite{Schonberger-CVPR16}. 
 The reconstruction process is the same as for LocalSfM, \ie, we first triangulate the 3D points and then use them to estimate the pose of the night-time query. 
 DenseSfM uses the same set of reference images for each query photo as LocalSfM.

\subsection{Localization from Multiple Images}
\PAR{Active Search + Generalized Cameras (Active Search+GC).} 
While most existing work on visual localization focuses on estimating the camera pose of an individual single query image, this paper additionally evaluates the benefits of using multiple images simultaneously for pose estimation. 
To this end, we assume that the relative poses between multiple query images are known and model these multiple images as a generalized camera. 
Given the matches found via Active Search for each individual image in a generalized camera, we use the 3-point-generalized-pose (GP3P) solver from \cite{Lee2015IJRR} to estimate the pose of the generalized camera. 
Together with the known relative poses, this provides us with a pose for each image in the generalized camera. 
We use these individual poses to evaluate the pose estimation accuracy. 
An inlier threshold of 12 pixels is used by RANSAC. 

Active Search+GC is not evaluated on the Aachen Day-Night dataset as it only provides individual query images. 
For the RobotCar Seasons, we evaluate two variants: 
\emph{Active Search+GC (triplet)} builds a generalized camera from images captured at the same point in time by the three cameras mounted on the RobotCar (left, rear, right). 
The resulting generalized cameras thus consist of three images each. 
\emph{Active Search+GC (sequence)} uses longer sequences taken with all three cameras. 
Each sequence consists of images taken consecutively in time under the same condition. 
More specifically, each sequence consists of a temporal sequence of images taken around the 49 manually selected reference positions (\cf Sec.~3.2 in the paper). 
For the CMU Seasons dataset, we only evaluate the Active Search+GC (sequence). All query images taken under the same condition for a given sub-model define one sequence.

In order to use the GP3P solver, the relative poses between the images in a generalized camera, as well as the scale of the relative translations between the images, need to be known. 
In our experiments, we extract the required relative poses directly from the ground truth camera poses. 
As a consequence, the results obtained with Active Search+GC (sequence) are optimistic in the sense that the method does not need to deal with the drift that normally occurs when estimating a trajectory via SLAM or SfM. 
Notice that we only use the relative poses. No information about the absolute pose of a generalized camera is used during pose estimation. 
Also, notice that the results obtained for Active Search+GC (triplet) are realistic: In this case, we are only using the known extrinsic calibration between the three cameras mounted on the RobotCar to define each generalized camera.

We also experimented with relative poses generated by our own multi-camera visual odometry (VO) system. 
Tab.~6 in the paper compares the results obtained when using ground truth poses with those obtained when using poses estimated by our VO pipeline on the night-time images of the RobotCar Seasons dataset. 
As can be seen, using ground truth poses leads to better results as generalized camera pose solvers are typically sensitive to calibration errors. 
Still, Active Search+GC with VO poses outperforms single image-based methods. 
We also evaluated Active Search+GC (sequence) on the CMU datasets, but found that the drift in the odometry was too severe to provide accurate camera poses. 
An interesting experiment would be to use only short sub-sequences (for which the drift is not too large) rather than the full sequences.



\begin{table}
\centering
\begin{tabular}{c|c}
\textbf{Parameter} & \textbf{Value} \\ \hline
Image Size & $48\times48$ ($144\times48$) \\ \hline
Patch Size & $8\times8$ \\ \hline
Sequence Length $d_s$ & 10
\end{tabular}
\vspace{2mm}
\caption{SeqSLAM parameters.}
\label{tab:seqslam-parameters}
\end{table}

\PAR{SeqSLAM.}
We used the OpenSeqSLAM implementation from \cite{sunderhauf2013we} with default parameters for template learning and trajectory uniqueness.
For each set of synchronized Grasshopper2 images, we downscale the original $1024\times1024$ resolution to $48\times48$, then concatenate all three images to form a $144\times48$ pixel composite.
The trajectory length parameter $d_s$ was set to 10 images; as both the query and database images are evenly spaced this corresponds to a trajectory length of 10 meters.
Tab.~\ref{tab:seqslam-parameters} summarizes the parameters used for the RobotCar experiments. 

\begin{table}[t]
\begin{center}
\setlength{\tabcolsep}{4pt}
\footnotesize{
\begin{tabular}{|l|c|c|c|c|}
\hline
 &  & \multicolumn{3}{c|}{\# images} \\
 condition & recorded & individual & triplets & sequences \\
\hline\hline
overcast (reference) & 28 Nov 2014 & 20,862 & 8,707 & - \\
\hline\hline
dawn & 16 Dec 2014 & 1,449 & 483 & 54 \\
dusk & 20 Feb 2015 & 1,182 & 394 & 48 \\
night & 10 Dec 2014 & 1,314 & 438 & 49 \\
night+rain & 17 Dec 2014 & 1,320 & 440 & 51 \\
overcast (summer) & 22 May 2015 & 1,389 & 463 & 52 \\
overcast (winter) & 13 Nov 2015 & 1,170 & 390 & 49 \\
rain & 25 Nov 2014 & 1,263 & 421 & 50 \\
snow & 3 Feb 2015 & 1,467 & 489 & 56 \\
sun & 10 Mar 2015 & 1,380 & 460 & 51 \\
\hline
total query & - & 11,934 & 3,978 & 460\\
\hline
\end{tabular}
}
\end{center}
\caption{Detailed statistics for the \emph{RobotCar Seasons} dataset. We used images from the \emph{overcast (reference)} traversal to build a 3D scene model. For each of the query sequences, we report the total number of query images taken by all three individual cameras, the resulting number of triplets used for Active Search+GC (triplet), and the number of temporally continuous query sequences used for Active Search+GC (sequence).}%
\label{tab:robotcar}%
\end{table}

\begin{table}[t]
\begin{center}
\setlength{\tabcolsep}{4pt}
\footnotesize{
\begin{tabular}{|l|c|c|c|}
\hline
 &  & \multicolumn{2}{c|}{\# images} \\
 condition & recorded & individual & sequences \\
\hline\hline
Sunny + No Foliage (reference) & 4 Apr 2011 & 7,159 & 17 \\
\hline\hline
Sunny + Foliage & 1 Sep 2010 & 8,076 & 16 \\
Sunny + Foliage & 15 Sep 2010 & 7,260 & 17 \\
Cloudy + Foliage & 1 Oct 2010 & 7,185 & 17 \\
Sunny + Foliage & 19 Oct 2010 & 6,737 & 17 \\
Overcast + Mixed Foliage & 28 Oct 2010 & 6,744 & 17\\
Low Sun + Mixed Foliage & 3 Nov 2010 & 6,982 & 17 \\
Low Sun + Mixed Foliage & 12 Nov 2010 & 7,262 & 17 \\
Cloudy + Mixed Foliage & 22 Nov 2010 & 6,649 & 17 \\
Low Sun + No Foliage + Snow & 21 Dec 2010 & 6,825 & 17\\
Low Sun + No Foliage & 4 Mar 2011 & 6,976 & 17\\
Overcast + Foliage & 28 Jul 2011 & 4,639 & 17\\
\hline
total query & - & 75,335 & 186\\
\hline
\end{tabular}
}
\end{center}
\caption{Detailed statistics for the \emph{CMU Seasons} dataset. We used images from the \emph{reference} traversal to build a 3D scene model. For each of the query sequences, we report the total number of query images taken and the number of temporally continuous query sequences used for Active Search+GC (sequence).}%
\label{tab:cmu}%
\end{table}


\begin{table}[t]
\begin{center}
\setlength{\tabcolsep}{4pt}
\footnotesize{
\begin{tabular}{|l|c|c|c|}
\hline
 Scene & Sub-model & \# images \\
\hline\hline
Urban & 1 - 7 &  31,250\\
Suburban & 8 - 10 & 13,736\\
Park & 11 - 17 & 30,349\\
\hline
total query & - & 75,335 \\
\hline
\end{tabular}
}
\end{center}
\caption{The type of scenery (urban, suburban and park) depicted in  the different sub-models of the \emph{CMU Seasons} dataset and the  total number of query images for each type. In total there are 31,250 urban, 13,736 suburban and 30,349 park images.}%
\label{tab:cmu2}%
\end{table}

\section{Dataset Details}
\label{sec:datasets}
\noindent This section provides additional details for the RobotCar Seasons and CMU Seasons dataset. More specifically, Tab.~\ref{tab:robotcar} details the time at which the individual traversals were recorded, the number of images per traversal, as well as the number of triplets and sequences used for Active Search+GC. 
Tab.~\ref{tab:cmu} provides similar details for the CMU Seasons dataset. In addition to listing the conditions for the different recordings, Tab.~\ref{tab:cmu2} lists the respective scenery (urban, suburban and park) for the different sub-models. 


\begin{table*}[th]
\begin{center}
\footnotesize{
\begin{tabular}{l|c|c|c|c|c|c|c}
 & \multicolumn{2}{c|}{Aachen Day-Night} & \multicolumn{4}{c|}{RobotCar Seasons} & CMU Seasons\\ \cline{2-8}
 &  & & \multicolumn{2}{c|}{Day} & \multicolumn{2}{c|}{Night} & \\ \cline{4-7}
Method & Day & Night & full model & sub-models & full model & sub-models & All \\ \hline
Active Search & 0.102 & 0.140 & 0.291 & 0.061 & 0.973 & 0.093 & 0.065 \\ \hline
CSL & 168.6 & 206.2 & 32.9 & $90.3^{\dagger}$ & 66.3 & $90.3^{\dagger}$ & 30.7\\ \hline
DenseVLAD \textbf{*} & 0.752 & 0.527 & 0.338 & - & 0.338 & - & 0.785\\ \hline
NetVLAD \textbf{$^\diamond$} & 0.105 & 0.105 & 0.137 & - & 0.137 & - & 0.107\\ \hline
FABMAP & 0.008 & 0.008 & 0.039 & - & 0.039 & - & 0.013\\ \hline \hline
Active Search+GC (triplet) & - &- & 0.879 & 0.180 & 2.940 & 0.289 & - \\ \hline
Active Search+GC (sequence) & - & - & 1.570 & 0.317 & 5.267 & 0.515 & 26.278 \\ \hline
seqSLAM & - & - & 0.251 & - & 0.251 & - & -\\ \hline \hline
LocalSfM\enspace \textbf{*},\textbf{$^\diamond$} & - & 19.591 & \multicolumn{2}{c|}{-} & \multicolumn{2}{c|}{22.486} & 44.577 \\ \hline
DenseSfM \textbf{*} & - & 16.719 & - & - & - & - & - \\ \hline
\end{tabular}
}
\end{center}
\caption{Average run-time per method on our three datasets. All timings are given in seconds. The timings include the time required for matching and (if applicable) spatial verification. Feature extraction times however are excluded. For Active Search+GC, which performs pose estimation using multiple cameras, run-times are typically dominated by the feature matching step (which is performed for each image that is part of a generalized camera). Methods marked with \textbf{*} are parallelized over multiple threads; all other methods utilize only a single CPU thread. Methods marked with a \textbf{$^\diamond$} symbol use the GPU, \eg, for feature matching. The two sub-model query times for the CSL are marked with $\dagger$ since the day and night queries were not timed separately, and the reported time is the average time per query over all queries (both day and night). } 
\label{tab:run-times}%
\end{table*}

\section{Timing Results}
\label{sec:timings}
\noindent Tab.~\ref{tab:run-times} provides an overview over the run-times of the various methods used for experimental evaluation on the three benchmark datasets. 
Timings are given in seconds and do include feature matching and (if applicable) camera pose estimation. However, feature extraction times are not included in the run-times since most algorithms are independent of the underlying feature representation and, in extension thereof, the specific implementation used to extract the features. 

We ran the different algorithms on different machines. For all variants of Active Search, a PC with an Intel Core i7-4770 CPU with 3.4GHz, 32GB of RAM, and an NVidia GeForce GTX 780 GPU was used. The same machine was used to run LocalSfM. Notice that due to their need to match multiple images, most of the run-time of Active Search+GC and LocalSfM is spent on feature matching. 
The increase in run-time for LocalSfM from the Aachen Day-Night to the RobotCar Seasons and CMU Seasons datasets is caused by the number of reference images considered for each dataset. 
For Aachen Day-Night, at most 20 reference images are considered per query while more images are used on the other two datasets (where more reference images are used for the CMU dataset due to a higher sampling density of the reference images). 
CSL was run on a computer cluster using an Intel Xeon E5-2650 v3 with 3.2 GB RAM per CPU core. 
The fact that CSL is substantially slower on the Aachen Day-Night dataset than on the RobotCar Seasons and CMU Seasons datasets is due to the image resolution of the query images. 
The query images of the Aachen Day-Night dataset have a higher resolution, which results in more detected local features. 
More features in turn lead to more matches and thus a significant increase in run-time for CSL due to its computational complexity of $\mathcal{O}(N^2 \log N)$ for $N$ matches. 
Both FAB-MAP and SeqSLAM results were generated using a single core of an Intel Core i7-4790K CPU with 4.0GHz and 32GB of RAM. 
DenseVLAD, NetVLAD\footnote{The run-time for NetVLAD includes the intermediate convolutional layer computation essentially corresponding to feature extraction.}, and DenseSfM were run on an Intel Xeon E5-2690 v4 with 2.60GHz with 256GB of RAM and an NVidia GeForce TitanX.

\section{Experimental Evaluation for All Conditions on CMU Seasons}
\label{sec:cmu}
\noindent  Due to space constraints, Sec.~6.3 of the paper only evaluates two types of conditions on the CMU dataset: Changes in foliage (foliage fully present, foliage somewhat present, no foliage) and differences in the type of scenery (urban, suburban, park) as these conditions are not covered by the other two datasets in our benchmark. 
Tab.~\ref{tab:full_cmu} provides the full evaluation of the different state-of-the-art algorithms on the CMU dataset. 

\begin{table*}[th!]
\begin{center}
\begin{minipage}{.5\linewidth}
    \scriptsize{

\setlength\tabcolsep{1.5pt}\begin{tabular}{l|c|c|c|}
& foliage & mixed foliage & no foliage\\ \cline{2-4}
\multicolumn{1}{r|}{\begin{tabular}[c]{@{}r@{}}m\\ deg\end{tabular}} & \begin{tabular}[c]{@{}c@{}}.25/.50/5.0\\ 2/5/10\end{tabular} & \begin{tabular}[c]{@{}c@{}}.25/.50/5.0\\ 2/5/10\end{tabular} & \begin{tabular}[c]{@{}c@{}}.25/.50/5.0\\ 2/5/10\end{tabular}\\ \hline
Active Search & 28.8 /\0 32.5 /\0 35.9 & 25.1 /\0 29.4 /\0 33.9 & 52.5 /\0 59.4 /\0 66.7 \\ \hline 
CSL & 16.3 /\0 19.1 /\0 26 & 15.2 /\0 18.8 /\0 28.6 & 36.5 /\0 43.2 /\0 57.5 \\ \hline 
DenseVLAD & 13.2 /\0 31.6 /\0 82.3 & 16.2 /\0 38.1 /\0 85.4 & 17.8 /\0 42.1 /\0 91.3 \\ \hline 
NetVLAD & 10.4 /\0 26.1 /\0 80.1 & 11.0 /\0 26.7 /\0 78.4 & 11.8 /\0 29.1 /\0 82 \\ \hline 
FABMAP & \01.1 /\0 2.7 /\0 16.5 & 1.0 /\0 2.5 /\0 14.7 & 3.6 /\0 7.9 /\0 30.7 \\ \hline 
LocalSfM & 55.4 /\0 57.0 /\0 59.9 & 52.4 /\0 55.1 /\0 58.6 & 70.8 /\0 72.7 /\0 75.9 \\ \hline 
AS+GC(seq) & 86.6 /\0 93.0 /\0 99.3 & 76.3 /\0 88.5 /\0 99.8 & 77.6 /\0 86.8 / 99.8 \\ \hline 
\end{tabular}
}%
\end{minipage}%
\begin{minipage}{.5\linewidth}
\scriptsize{

\setlength\tabcolsep{1.5pt}\begin{tabular}{l|c|c|c|}
& urban & suburban & park\\ \cline{2-4}
\multicolumn{1}{r|}{\begin{tabular}[c]{@{}r@{}}m\\ deg\end{tabular}} & \begin{tabular}[c]{@{}c@{}}.25/.50/5.0\\ 2/5/10\end{tabular} & \begin{tabular}[c]{@{}c@{}}.25/.50/5.0\\ 2/5/10\end{tabular} & \begin{tabular}[c]{@{}c@{}}.25/.50/5.0\\ 2/5/10\end{tabular}\\ \hline
Active Search & 55.2 /\0 60.3 /\0 65.1 & 20.7 /\0 25.9 /\0 29.9 & 12.7 /\0 16.3 /\0 20.8 \\ \hline 
CSL & 36.7 /\0 42.0 /\0 53.1 & 8.6 /\0 11.7 /\0 21.1 & 7 /\0 9.6 /\0 17.0 \\ \hline 
DenseVLAD & 22.2 /\0 48.7 /\0 92.8 & 9.9 /\0 26.6 /\0 85.2 & 10.3 /\0 27.0 /\0 77.0 \\ \hline 
NetVLAD & 17.4 /\0 40.3 /\0 93.2 & 7.7 /\0 21.0 /\0 80.5 & 5.6 /\0 15.7 /\0 65.8 \\ \hline 
FABMAP & 2.7 /\0 6.4 /\0 27.3 & 0.5 /\0 1.5 /\0 13.6 & 0.8 /\0 1.7 /\0 11.5 \\ \hline 
LocalSfM & 72.8 /\0 74.1 /\0 76.1 & 55.2 /\0 57.7 /\0 61.3 & 41.8 /\0 44.5 /\0 48.7 \\ \hline 
AS+GC(seq) & 86.4 /\0 93.6 /\0 99.8 & 92.0 /\0 96.0 /\0 99.7 & 71.0 /\0 84.0 /\0 99.2 \\ \hline 
\end{tabular}
}
\end{minipage} 
\vspace{0.5cm}
\scriptsize{

\setlength\tabcolsep{1.5pt}\begin{tabular}{l|c|c|c|c|c|}
& sunny & low-sun & cloudy & overcast & snow \\ \cline{2-6}
\multicolumn{1}{r|}{\begin{tabular}[c]{@{}r@{}}m\\ deg\end{tabular}} & \begin{tabular}[c]{@{}c@{}}.25/.50/5.0\\ 2/5/10\end{tabular} & \begin{tabular}[c]{@{}c@{}}.25/.50/5.0\\ 2/5/10\end{tabular} & \begin{tabular}[c]{@{}c@{}}.25/.50/5.0\\ 2/5/10\end{tabular} & \begin{tabular}[c]{@{}c@{}}.25/.50/5.0\\ 2/5/10\end{tabular} & \begin{tabular}[c]{@{}c@{}}.25/.50/5.0\\ 2/5/10\end{tabular}\\ \hline
Active Search & 27.3 /\0 30.9 /\0 34.1 & 34.6 /\0 39.8 /\0 45.5 & 34.6 /\0 39.4 /\0 44.0 & 30.0 /\0 34.3 /\0 38.4 & 41.4 /\0 49.0 /\0 57.5 \\ \hline 
CSL & 15.5 /\0 18.3 /\0 24.8 & 22.6 /\0 27.4 /\0 38.8 & 21.7 /\0 25.8 /\0 35.8 & 17.6 /\0 20.8 /\0 29.2 & 26.0 /\0 33.2 /\0 49.1 \\ \hline 
DenseVLAD & 13.2 /\0 31.3 /\0 81.4 & 15.1 /\0 36.9 /\0 86.0 & 18.5 /\0 41.9 /\0 89.0 & 15.1 /\0 35.2 /\0 85.2 & 17.4 /\0 41.3 /\0 87.2 \\ \hline
NetVLAD & 10.5 /\0 25.9 /\0 79.2 & 10.1 /\0 25.7 /\0 77.7 & 13.0 /\0 30.5 /\0 82.9 & 10.9 /\0 27.0 /\0 82.7 & 10.2 /\0 25.2 /\0 75.5 \\ \hline 
FABMAP & 1.0 /\0 2.5 /\0 15.2 & 2.0 /\0 4.6 /\0 20.8 & 1.8 /\0 4.1 /\0 20.1 & 0.9 /\0 2.7 /\0 17.0 & 2.2 /\0 4.8 /\0 22.4 \\ \hline 
LocalSfM & 53.5 /\0 55.1 /\0 58.0 & 56.9 /\0 59.5 /\0 62.8 & 63.5 /\0 65.3 /\0 68.4 & 57.0 /\0 59.1 /\0 62.4 & 64.2 /\0 66.6 /\0 70.4 \\ \hline 
AS+GC(seq) & 85.8 /\0 92.1 /\0 99.5 & 75.5 /\0 86.3 /\0 99.8 & 81.3 /\0 92.7 /\0 99.8 & 86.2 /\0 93.0 /\0 98.8 & 69.4 /\0 76.5 /\0 99.8 \\ \hline 
\end{tabular}
}
\end{center}
\caption{Full evaluation on the \textbf{CMU Seasons} dataset. Besides evaluating the impact of foliage (top left) and the type of environment (top right) on the pose estimation accuracy, both of which were already presented in the main paper, we also evaluate the impact of different weather conditions.}%
\label{tab:full_cmu}%
\end{table*}

\section{Cumulative Distributions of Position and Orientation Errors}
\label{sec:distributions}
\begin{figure*}[tbp]
\centering
\includegraphics[width=0.98\linewidth]{\Figssupp/legend}
\begin{tabular}{@{\hskip 2pt}c@{\hskip 0pt}c@{\hskip 0pt}c@{\hskip 0pt}c@{\hskip 0pt}c@{\hskip 2pt}}
\includegraphics[height=2.6cm]{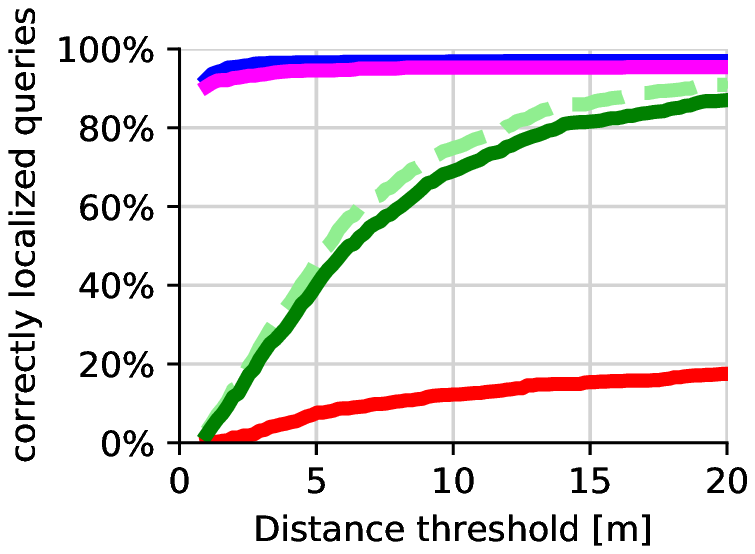} &
\includegraphics[height=2.6cm]{\Figssupp/Aachen_night_distance_new} &
\includegraphics[height=2.6cm]{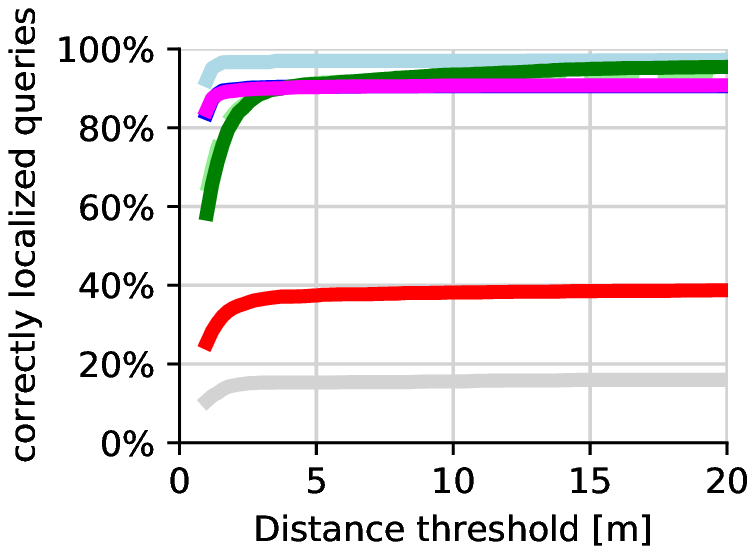} &
\includegraphics[height=2.6cm]{\Figssupp/RobotCar_Rear_night_distance_new} &
\includegraphics[height=2.6cm]{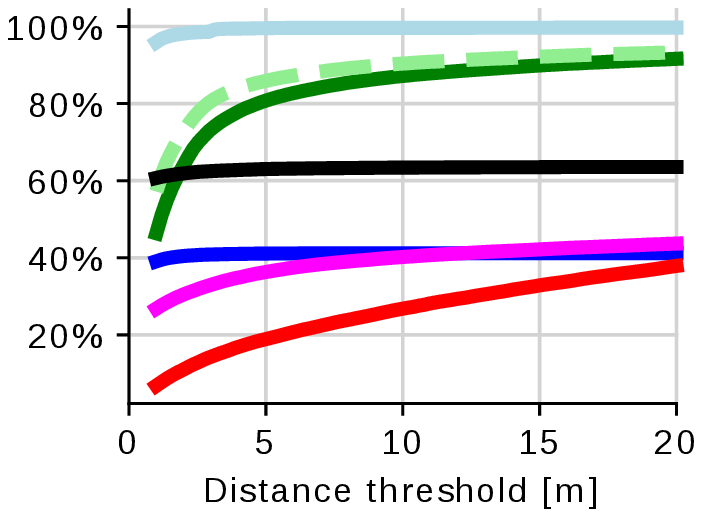} \\
\end{tabular}
\begin{tabular}{@{\hskip 2pt}c@{\hskip 0pt}c@{\hskip 0pt}c@{\hskip 0pt}c@{\hskip 0pt}c@{\hskip 2pt}}
\includegraphics[height=2.6cm]{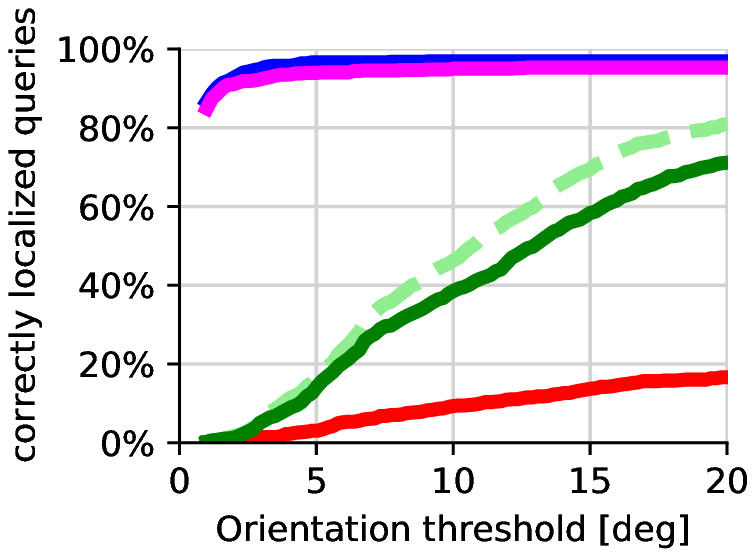} &
\includegraphics[height=2.6cm]{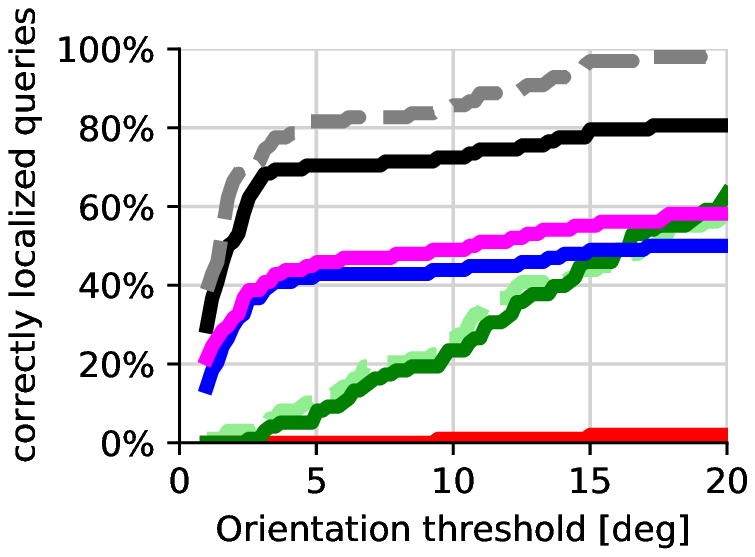} &
\includegraphics[height=2.6cm]{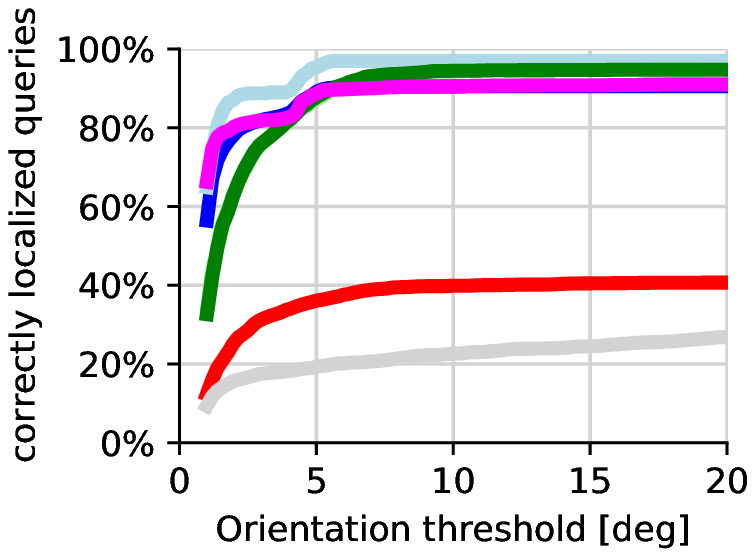} &
\includegraphics[height=2.6cm]{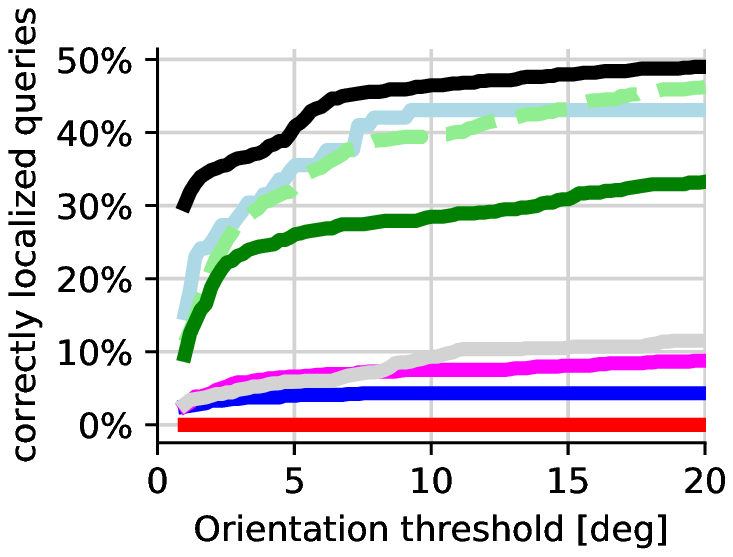} &
\includegraphics[height=2.6cm]{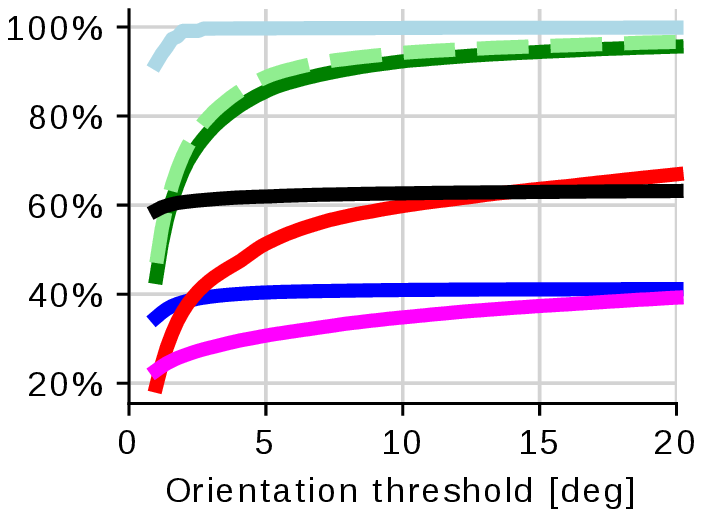} \\
{\small (a) Aachen - day} & {\small (b) Aachen - night} & {\small (c) RobotCar - day} & {\small (d) RobotCar - night} & {\small (e) CMU - all}\\
\end{tabular}
\vspace{1mm}
\caption{Cumulative distribution of position and orientation errors for the three datasets.
}
\label{fig:results:distributions}
\end{figure*}

\noindent Fig.~4 in the paper shows the cumulative distributions of the position errors of the evaluated methods for the night-time queries of the Aachen Day-Night and RobotCar Seasons datasets. 
For completeness, Fig.~\ref{fig:results:distributions} shows cumulative distributions of the position and orientation errors for all datasets. 
Notice that the results reported in the tables in the paper and the appendix are based on thresholding both the position \emph{and} orientation error. 
Thus, the percentage of localized query images reported in the table is lower than the curves shown in Fig.~\ref{fig:results:distributions}, which are obtained by thresholding either the position or orientation error.

{\small
\bibliographystyle{ieee}
\bibliography{vgg_bib/shortstrings,vgg_bib/vgg_local,vgg_bib/vgg_other,local}
}

\end{document}